\documentclass{article} %
\usepackage{iclr2025_conference,times}

\usepackage{amsmath,amsfonts,bm}

\def\eqref#1{equation~\ref{#1}}

\def\1{\bm{1}}

\DeclareMathAlphabet{\mathsfit}{\encodingdefault}{\sfdefault}{m}{sl}
\SetMathAlphabet{\mathsfit}{bold}{\encodingdefault}{\sfdefault}{bx}{n}

\usepackage{hyperref}
\usepackage{url}
\usepackage{subcaption}
\usepackage{nicematrix}
\usepackage{xcolor}
\usepackage{colortbl}
\usepackage{booktabs}
\usepackage{algorithmic}
\usepackage{multirow}  %
\usepackage{graphicx}  %
\usepackage{listings}
\usepackage{wrapfig}
\usepackage{tcolorbox}

\usepackage[utf8]{inputenc}  %
\usepackage{amsmath,amssymb,booktabs}

\usepackage{upquote}        %
\usepackage{caption}
\usepackage[T1]{fontenc}
\usepackage{inconsolata}    %

\usepackage[frozencache,cachedir=minted-cache]{minted}
\setminted{
  fontsize=\small,
  breaklines,
  linenos,
  frame=single,
  framesep=3mm,
  bgcolor=black!2
}

\title{Confidence-Guided Self-Refinement}

\iclrfinalcopy

\author{
Chen Jin$^{1}$, \
Ryutaro Tanno$^{2}$, \
Tom Diethe$^1$, \
Philip Teare$^{1}$\\
$^1$Centre for AI, AstraZeneca, Cambridge, UK\\
$^2$Google DeepMind, UK\\
}

\newcommand{\std}[1]{\textsubscript{\textcolor{gray}{\scriptsize±#1}}}

\usepackage[ruled,vlined]{algorithm2e}
\definecolor{commentcolor}{RGB}{110,154,155}   %
\newcommand{\PyComment}[1]{\ttfamily\textcolor{commentcolor}{\# #1}}  %

\begin{document}

\maketitle
\footnotetext[1]{Email: \texttt{chen.jin@astrazeneca.com}}

\begin{abstract}
Large Language Models (LLMs) often rely on test-time scaling via parallel decoding (e.g., 512 samples) to boost reasoning accuracy, but this incurs substantial compute. 
We introduce \textbf{CoRefine}, a confidence-guided self-refinement method that achieves competitive accuracy at a fraction of the tokens via a lightweight \(\sim 211\mathrm{k}\)-parameter Conv1D controller atop a frozen LLM.
The controller consumes \emph{full-trace confidence} to decide whether to halt, re-examine, or try a different approach—enabling targeted self-correction with an average of \(\sim 2.7\) refinement steps per problem (\(\approx 190\times\) token reduction) relative to 512-sample baselines.
Across diverse reasoning benchmarks and three open-source models, the controller achieves 92.6\% precision when it confidently halts, indicating that confidence dynamics reliably signal correctness without ground-truth verification.
We extend this to \textbf{CoRefine-Tree}, a hybrid sequential–parallel variant that adaptively balances exploration and exploitation, with easy serving integration and verifier compatibility.
By treating confidence as a \emph{control signal} rather than a correctness guarantee, CoRefine provides a modular primitive for scalable reasoning and agentic settings with imperfect verifiers.
\end{abstract}

\begin{figure}[h!]
\centering
\includegraphics[width=1.0\textwidth]{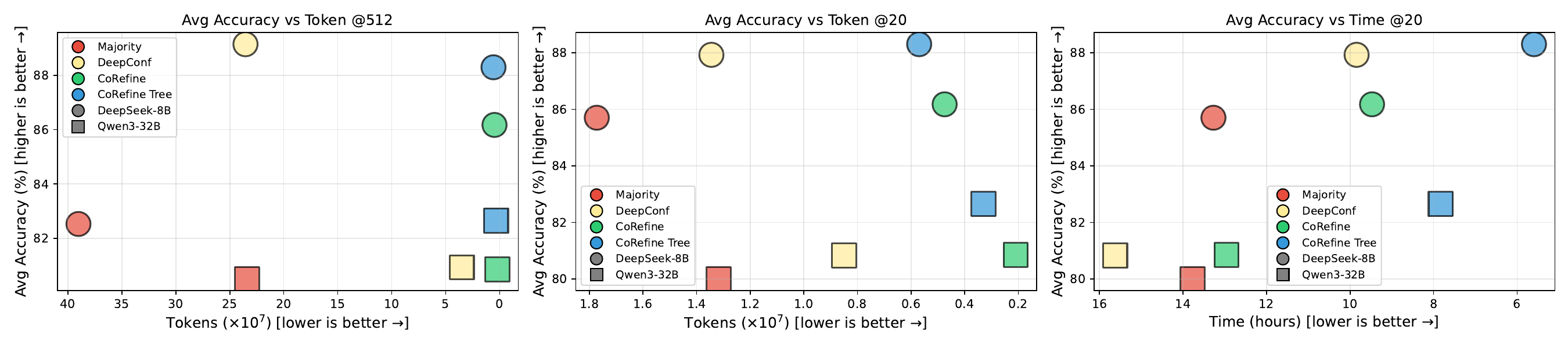}
\centering
\includegraphics[width=0.9\textwidth]{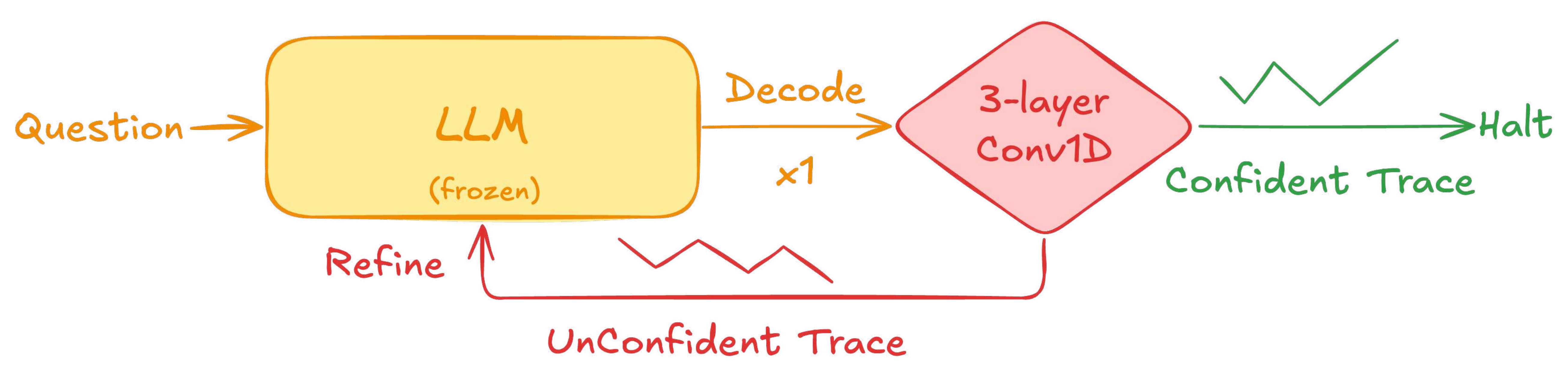}
\caption{\textbf{Top:} Token efficiency versus accuracy across four reasoning benchmarks: AIME24, AIME25, BRUMO25 and HMMT25. CoRefine achieves competitive or superior accuracy to 512-sample or 20-sample majority voting with $\sim$190$\times$ fewer tokens. Wall-clock time versus accuracy showing that token savings translate to actual latency reduction, with CoRefine saving up to 63\% over parallel baselines. \textbf{Bottom:} Confidence-Guided Self-Refine overview. The controller consumes full-trace confidence features of the LLM decoded reasoning chain and decides: HALT (accept current answer), RETHINK (verify reasoning), or ALTERNATIVE (explore new approach).}
\label{fig:highlight}
\end{figure}

\section{Introduction}

Test-time scaling improves LLM reasoning accuracy but incurs substantial compute~\citep{snell2024scaling,welleck2024decoding}. Parallel methods like \emph{self-consistency}~\citep{Wang2023SelfConsistency} sample hundreds of reasoning paths and aggregate via majority voting, trading compute for accuracy at proportionally scaling costs. To illustrate: achieving a 14-point accuracy improvement on AIME 2025 (from 68\% to 82\%) with DeepSeek-8B requires 512 parallel traces per problem, consuming over 100 million additional tokens~\citep{fu2025deepconf}.

An alternative paradigm is \emph{sequential refinement}, where the model iteratively improves its answer based on feedback~\citep{madaan2023self}. However, refinement suffers from compounding errors---early mistakes amplify downstream~\citep{Sang2025AutoCrit}---and existing approaches face two fundamental challenges: (1) \textbf{when to stop}---models often lack principled criteria for halting refinement, leading to over-iteration or premature termination~\citep{Seo2024Rethinking}; and (2) \textbf{how to refine}---generic ``rethink'' prompts may not provide sufficient guidance for targeted improvement. Indeed, recent studies show that refined outputs are not always superior to original versions~\citep{Seo2024Rethinking}, and that LLMs fail to correct errors over half the time without external guidance~\citep{Feng2025Misinfo}.

We propose \textbf{CoRefine: Confidence-Guided Self-Refinement}, which addresses both challenges by using \emph{confidence as a control signal}. The key insight is that token-level prediction confidence, aggregated across a reasoning trace, provides a rich signal for deciding whether to accept the current answer (HALT), re-examine it (RETHINK), or try a different approach (ALTERNATIVE). This framing treats refinement as an \emph{exploration-exploitation tradeoff}~\citep{Tang2024REx}, and crucially, uses confidence for \emph{adaptive compute allocation} rather than as a direct correctness estimate---even imperfectly calibrated confidence can guide useful refinement decisions (Section~\ref{sec:confidence-control}).

CoRefine consists of three components: (1) \textbf{full-trace confidence extraction} from the model's logprobs, (2) a \textbf{lightweight neural controller} ($\sim$211K parameters) that maps confidence features to refinement decisions, and (3) \textbf{targeted synthesis prompts} that compact previous reasoning into high-signal context for self-correction. The controller is trained via supervised learning on historical trajectories to predict oracle-optimal actions from confidence patterns. We further extend this to \textbf{CoRefine Tree}, a hybrid sequential--parallel variant that combines the token efficiency of sequential refinement with the robustness of parallel sampling.

Our approach offers several advantages:
\begin{itemize}
    \item \textbf{Efficiency:} CoRefine achieves parity or better accuracy than 512-sample majority voting with an average of $\sim$2.7 iterations, representing a $\approx$190$\times$ reduction in token usage, translating to 63\% wall-clock saving (Figure~\ref{fig:highlight}). 
    \item \textbf{Modularity:} The controller is a separate, frozen-LLM-compatible module that requires no backbone fine-tuning and integrates seamlessly with existing serving stacks.
    \item \textbf{Adaptivity:} The controller learns problem-specific stopping criteria—halting early on confident, consistent answers while allowing more exploration on difficult problems.
    \item \textbf{Reliability:} When the controller confidently decides to halt (majority of traces vote HALT), it achieves \textbf{92.6\% precision}—validating that confidence patterns provide a reliable signal for knowing \emph{when} the model has found the correct answer.
    \item \textbf{Adaptability to regulated domains:} By extending to a 4-class controller with REFUSE, CoRefine learns to distinguish genuine uncertainty from post-trained conservative behavior—predicting when encouragement will recover correct answers versus when honest abstention is appropriate (Section~\ref{sec:bixbench}).
    \item \textbf{Foundation for agentic systems:} By framing confidence as a control signal, CoRefine provides a principled primitive for future multi-agent systems where individual agents may have imperfect verifiers. Recent work on agent debugging shows that systematic learning from failures can improve task success rates by up to 26\%~\citep{Liang2025COCO}; CoRefine's ALTERNATIVE action provides a mechanism for such recovery.
\end{itemize}

We evaluate CoRefine across diverse mathematical reasoning benchmarks (AIME 2024/2025, HMMT 2025, BRUMO25) and multiple open-source models (DeepSeek-8B, Qwen3-32B). Results demonstrate that CoRefine consistently matches or outperforms parallel approaches while using orders of magnitude fewer computational resources.

\section{Confidence as a Control Signal}
\label{sec:confidence-control}

Before describing our method, we establish the empirical foundation: why confidence from token-level logprobs provides useful signal for refinement decisions, and how we can leverage it without assuming it directly estimates correctness.

\subsection{Token-Level Confidence Extraction}

Given a language model's predicted token distribution $P_i$ at position $i$, we compute \textit{token confidence} $C_i$ as the negative average log-probability of the top-$k$ tokens:
\begin{equation}
    C_i = - \frac{1}{k} \sum_{j=1}^{k} \log P_i(j),
\end{equation}
where $k$ denotes the number of top tokens considered (we use $k=20$). High confidence corresponds to peaked distributions and greater model certainty, while low confidence indicates uncertainty in token prediction.

For a complete reasoning trace of $N$ tokens, we aggregate these into a \textit{confidence trace} $\mathbf{c} = (C_1, C_2, \ldots, C_N)$. This full-trace representation captures the model's confidence dynamics throughout the reasoning process.

\subsection{Confidence Distributions: Correct vs. Incorrect}

\begin{figure}[t!]
\centering
\begin{subfigure}[b]{0.49\textwidth}
\centering
\includegraphics[width=\textwidth]{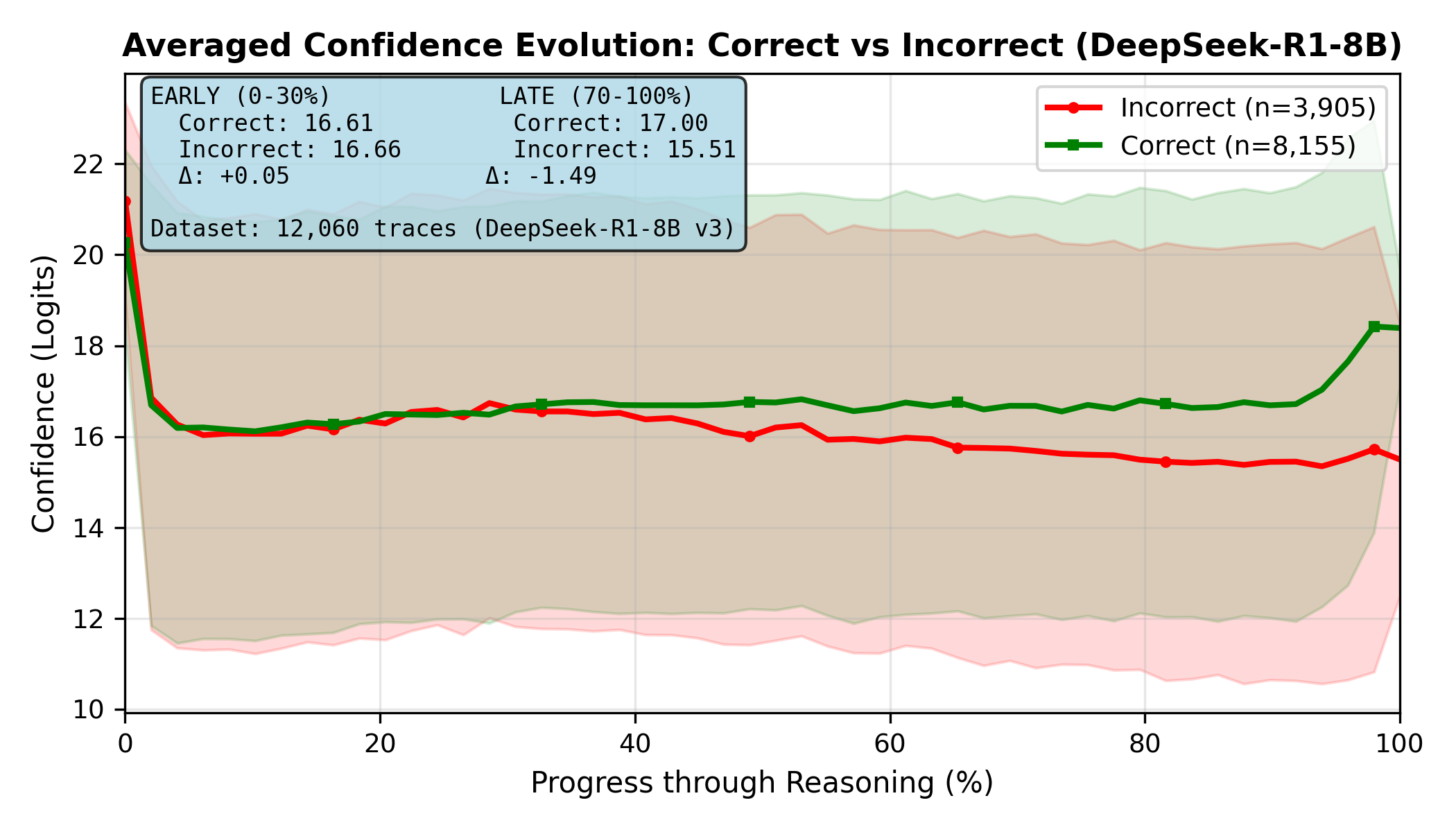}
\end{subfigure}
\hfill
\begin{subfigure}[b]{0.49\textwidth}
\centering
\includegraphics[width=\textwidth]{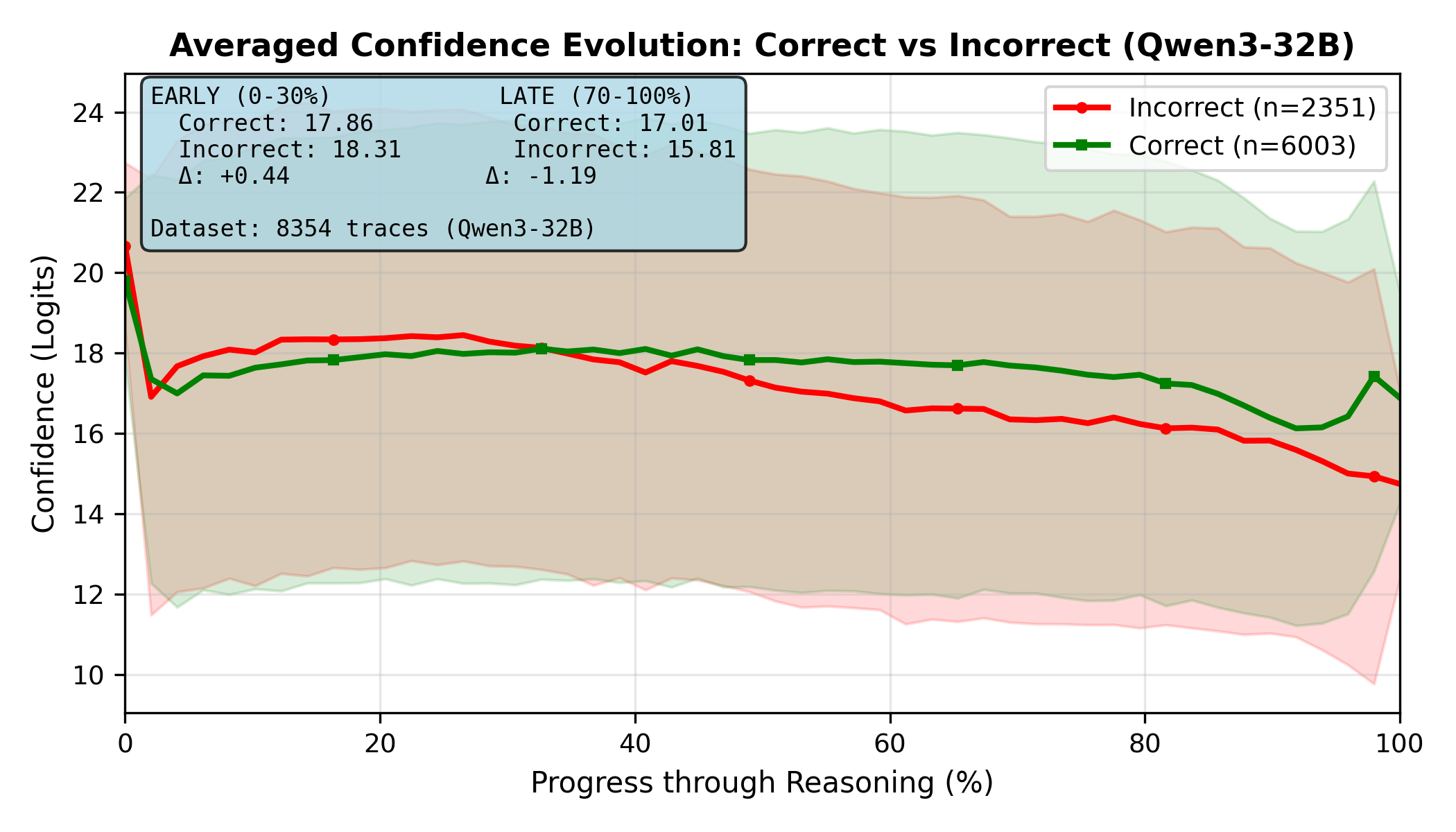}
\end{subfigure}
\caption{Averaged confidence evolution for correct vs. incorrect reasoning traces. \textbf{Left:} DeepSeek-R1-8B (12,060 traces). \textbf{Right:} Qwen3-32B (8,354 traces). Both models show correct traces maintaining higher late-phase confidence, but with distinct dynamics: DeepSeek exhibits increasing confidence for correct traces with a sharp terminal spike, while Qwen3 shows globally descending confidence for both classes.}
\label{fig:confidence_distributions}
\end{figure}

Figure~\ref{fig:confidence_distributions} shows confidence distributions for correct and incorrect reasoning traces. Analysis of 12,060 traces (DeepSeek-R1-8B: 8,155 correct, 3,905 incorrect) and 8,354 traces (Qwen3-32B: 6,003 correct, 2,351 incorrect) reveals:

\begin{enumerate}
    \item \textbf{Late-phase divergence:} Both models show correct traces achieving higher confidence in late phases (70-100\%). DeepSeek shows a $\Delta$=-1.49 gap (17.00 vs 15.51 logits) while Qwen3 shows $\Delta$=-1.19 (17.01 vs 15.81 logits), with correct traces consistently higher.
    \item \textbf{Early overconfidence paradox:} Counterintuitively, incorrect traces start with \emph{higher} confidence in early phases (0-30\%): DeepSeek shows $\Delta$=+0.05 (16.66 vs 16.61) and Qwen3 shows $\Delta$=+0.44 (18.31 vs 17.86). This suggests early confidence is misleading.
    \item \textbf{Model-specific dynamics:} DeepSeek exhibits increasing confidence for correct traces with a sharp ``hook'' spike at 95-100\%, while incorrect traces remain flat. Qwen3 shows globally descending confidence for both classes, but with steeper decline for incorrect traces. These distinct patterns motivate learning-based controllers over hand-crafted rules.
\end{enumerate}

\paragraph{Key insight: Control vs. Estimation.} These observations suggest that confidence is \emph{not} a reliable correctness estimator. However, it can still be a useful \emph{control signal}. Consider an analogy: a robot's wheel encoders may not perfectly measure position, but the robot can still use encoder deltas to decide when to turn. Similarly, confidence may not tell us if an answer is correct, but confidence \emph{patterns}—drops, stability, trends—can inform when to continue refining.

\subsection{From Confidence to Control Actions}

We define three control actions based on the controller's assessment of the confidence trace:

\begin{itemize}
    \item \textbf{HALT}: Accept the current answer. Triggered when confidence is stable and answer is consistent across iterations.
    \item \textbf{RETHINK}: Re-examine the reasoning. Triggered when confidence suggests potential errors but the overall approach seems sound.
    \item \textbf{ALTERNATIVE}: Try a completely different approach. Triggered when low confidence with inconsistent answers suggests the current path is unproductive.
\end{itemize}

The controller learns to map confidence features to these actions through supervised learning on trajectories where we know the eventual outcome (correct/incorrect). Importantly, the controller does not predict correctness directly—it predicts which \emph{action} is most likely to lead to a correct final answer.

\section{Confidence-Guided Self-Refine (CoRefine)}
\label{sec:method}

We now present CoRefine, our confidence-guided self-refinement framework. The system consists of three components: confidence feature extraction, a neural controller, and synthesis prompts for refinement.

\subsection{System Overview}

\begin{figure}[t]
    \centering
    \includegraphics[width=\textwidth]{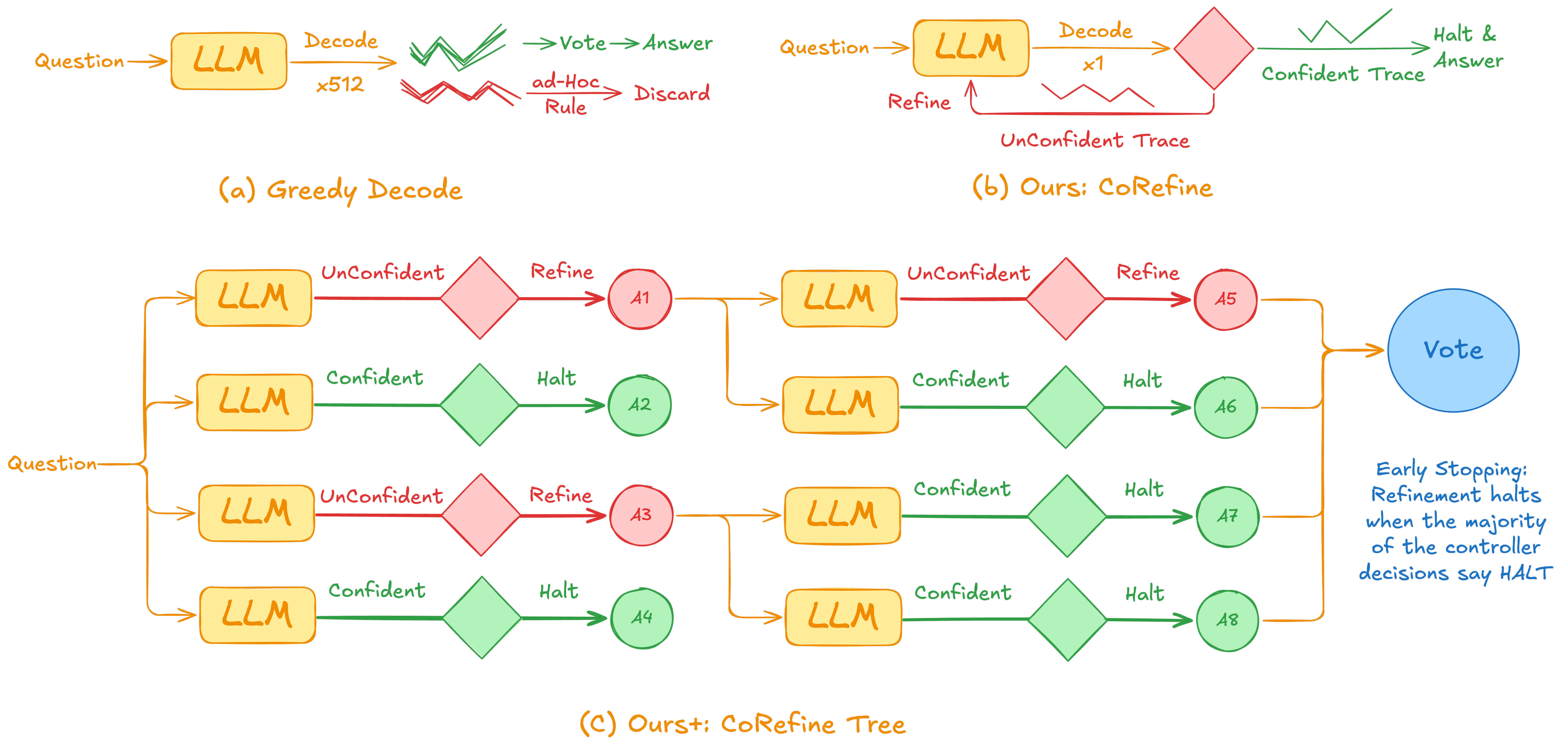}
    \caption{\textbf{(a) DeepConf - Parallel:} Sample $K$ traces, filter by confidence, aggregate via weighted voting. \textbf{(b) CoRefine - Sequential:} Iteratively refine using controller decisions based on full-trace confidence. \textbf{(c) CoRefine Tree - Hybrid:} Combine parallel sampling with sequential refinement for best of both paradigms.}
    \label{fig:DeepConf_vs_CoRefine}
\end{figure}

Figure~\ref{fig:DeepConf_vs_CoRefine} contrasts CoRefine with parallel (DeepConf~\citep{fu2025deepconf}) and hybrid approaches. CoRefine operates through an iterative refinement loop. At each iteration $t$, the system generates a response $y_t$ from the language model while simultaneously extracting the token-level confidence trace $\mathbf{c}_t$ from the model's logprobs. The final answer $a_t$ is then parsed from the response using standard extraction patterns (e.g., $\backslash$boxed\{\} notation for mathematical problems). These signals, together with the refinement history $h_t$, are transformed into a feature vector $\phi(\mathbf{c}_t, h_t)$ that captures both the current confidence dynamics and the trajectory of previous attempts.

The neural controller $\pi_\theta$ consumes this feature vector and outputs a probability distribution over three discrete actions. If the controller selects HALT, the system terminates and returns the current answer $a_t$. Otherwise, the system constructs a synthesis prompt incorporating the action type (RETHINK or ALTERNATIVE) and compacted summaries of previous reasoning attempts, then continues to the next iteration. This loop proceeds until either the controller issues a HALT decision or a maximum iteration budget is exhausted.

\subsection{Confidence Feature Extraction}

Given a confidence trace $\mathbf{c} = (C_1, \ldots, C_N)$ with $N$ tokens, we extract features designed to capture patterns useful for control decisions:

\paragraph{Temporal Downsampling.} Long traces (often 5,000+ tokens for mathematical reasoning) are downsampled to a fixed length $L$ using average pooling:
\begin{equation}
    \bar{c}_j = \frac{1}{|B_j|} \sum_{i \in B_j} C_i, \quad j = 1, \ldots, L
\end{equation}
where $B_j$ is the $j$-th bin of tokens. This aggressive downsampling (from $N \approx 5{,}000$--$20{,}000$ tokens to $L=16$ bins) is deliberate: as Figure~\ref{fig:confidence_distributions} shows, raw confidence traces exhibit substantial high-frequency noise, with token-level fluctuations obscuring the underlying correctness signal. Early experiments with minimal or no downsampling ($L=64$, $L=256$, or full-length traces) yielded worse controller accuracy, suggesting the noise overwhelms pattern detection. Average pooling acts as a low-pass filter, smoothing local variations while preserving the macro-level dynamics (early overconfidence, late-phase divergence) that discriminate correct from incorrect traces. The resulting feature vector $\phi_t = \bar{\mathbf{c}}_t \in \mathbb{R}^{16}$ serves as input to the controller. We also explored augmenting this representation with regional statistics and cross-iteration dynamics, but these provided marginal gains ($<$1\% accuracy); see Appendix~\ref{app:variants}.

\paragraph{Why Not Text Features?} A natural alternative would be to use the actual reasoning text (CoT) as controller input. We deliberately avoid this for two reasons. First, \textbf{computational cost}: processing long reasoning traces (5,000--20,000 tokens) would require a text encoder and likely fine-tuning, negating our goal of a lightweight, modular controller. Second, \textbf{noisy signal}: our early experiments with text-based uncertainty detection proved unreliable---hedging language (``maybe'', ``perhaps'', ``I think'') appeared frequently in \emph{correct} traces as well as incorrect ones, providing little discriminative power. In contrast, token-level confidence traces offer a compact, semantics-free signal that captures model uncertainty without parsing ambiguous linguistic cues.

\subsection{Neural Controller Architecture}

The controller $\pi_\theta: \mathbb{R}^{16} \rightarrow \Delta^3$ maps the downsampled confidence trace to a probability distribution over three actions. We employ a one-dimensional convolutional architecture that naturally captures temporal patterns:
\begin{equation}
\begin{aligned}
    h^{(1)} &= \text{ReLU}(\text{Conv1D}_{64}(\bar{\mathbf{c}})), \quad
    h^{(2)} = \text{ReLU}(\text{Conv1D}_{128}(h^{(1)})), \\
    h^{(3)} &= \text{ReLU}(\text{Conv1D}_{256}(h^{(2)})), \quad
    \pi_\theta(\phi) = \text{Softmax}(\text{MLP}(\text{Flatten}(h^{(3)})))
\end{aligned}
\end{equation}
The convolutional layers employ kernel sizes $[5, 5, 3]$ with stride 2, enabling hierarchical extraction of local confidence fluctuations into progressively abstract representations. We chose Conv1D over fully-connected (MLP) architectures based on empirical comparison: MLPs achieved 78--80\% validation accuracy versus Conv1D's 83--84\%, likely because Conv1D provides translation invariance---the same confidence pattern (e.g., a mid-trace dip followed by recovery) is detected regardless of its absolute position. This property is crucial since diagnostic patterns can occur at varying points in a reasoning trace. The architecture maintains extreme parameter efficiency at approximately 211K parameters.

\paragraph{Training.} We train the controller via supervised learning on historical refinement trajectories. Given a dataset of $N_\text{train}$ training and $N_\text{val}$ validation traces from math problems (see Section~\ref{sec:experiments} for model-specific details), each labeled with oracle actions $(x_i, \{(y_t^{(i)}, \mathbf{c}_t^{(i)}, o_t^{(i)})\}_{t=1}^{T_i})$ where $o_t^{(i)} \in \{0,1,2\}$ denotes HALT, RETHINK, or ALTERNATIVE, we minimize cross-entropy loss augmented with a step penalty:
\begin{equation}
    \mathcal{L}(\theta) = \mathbb{E}_{i,t}\left[ -\log \pi_\theta^{o_t^{(i)}}(\phi_t^{(i)}) \right] + \lambda \cdot t
\end{equation}
The step cost $\lambda \cdot t$ (with $\lambda = 0.1$) encourages early halting when appropriate. Training uses Adam optimizer with learning rate $10^{-3}$, batch size 32, for 30 epochs. The controller converges to 83--84\% validation accuracy, demonstrating that raw confidence patterns alone provide sufficient signal for refinement decisions.

\paragraph{Oracle Label Generation.} Training data combines traces from two sources: (1) \emph{parallel sampling}, where multiple independent traces are generated per problem without iteration ($t=0$), and (2) \emph{sequential refinement}, where traces are generated through iterative runs ($t \geq 0$). Correct traces receive HALT labels regardless of source. For incorrect traces from sequential runs, we use iteration history: if a subsequent iteration eventually succeeds within the same approach, the label is RETHINK; if success requires a fundamentally different method, the label is ALTERNATIVE. For incorrect traces from parallel runs (no iteration history), we apply confidence-based heuristics: declining trend suggests RETHINK (foundational errors), stable early confidence with late drop suggests RETHINK (calculation errors), and high volatility suggests ALTERNATIVE (unstable approach requiring a fresh start).

\paragraph{Theoretical Justification.} A potential concern is circularity: for parallel traces without iteration history, labels are derived from confidence-based heuristics, yet the controller is trained to predict actions from confidence. We provide a Bayesian decision-theoretic analysis in Appendix~\ref{app:bayesian-justification} showing this does not introduce degeneracy. The key insight is that labels encode \emph{correctness} (ground-truth verification), not confidence---the heuristics merely approximate the counterfactual ``which action would have succeeded?'' for traces where we lack iteration history. Formally, the controller learns $P(a^* | \mathbf{c})$ where the optimal action $a^*$ is defined by correctness outcomes, and confidence $\mathbf{c}$ serves as a \emph{sufficient statistic} for predicting these outcomes. Three factors break potential circularity: (1) correct traces (67--72\% of data) receive HALT labels based purely on ground-truth verification; (2) sequential traces provide causal ground truth from actual iteration outcomes; and (3) the heuristics encode domain knowledge about error types (not just pattern matching), serving as an informative prior that the controller refines through supervised learning.

\subsection{Synthesis Prompts for Refinement}

When the controller selects a refinement action, we construct a \emph{synthesis prompt} that provides the language model with structured context for its next attempt. The prompt integrates four components: the original problem statement, compacted summaries of previous reasoning attempts, the specific action type (RETHINK or ALTERNATIVE), and aggregate confidence statistics from prior iterations.

\paragraph{Message Compaction.} Long-form reasoning traces (often exceeding 10,000 tokens) must be compressed to fit within context limits while preserving actionable information. Our heuristic compaction extracts the final answer (parsed from $\backslash$boxed\{\} notation) and key intermediate steps. This compacted representation typically reduces trace length by 90--95\% while retaining the information most relevant for subsequent refinement.

\paragraph{Action-Specific Instructions.} The synthesis prompt includes action-specific guidance that shapes the model's refinement strategy. For RETHINK actions, we provide a truncated window of the previous reasoning (approximately 800 tokens from the end) and instruct the model to review its approach, identify weak points, and check for calculation errors or logical gaps---encouraging careful reconsideration rather than wholesale abandonment. For ALTERNATIVE actions, we explicitly direct the model to try a completely different method or problem formulation, signaling that the previous approach may have fundamental issues that cannot be resolved through incremental correction.

Algorithm~\ref{alg:corefine} summarizes the complete CoRefine procedure, showing how confidence extraction, controller decisions, and synthesis prompts integrate into the iterative refinement loop.

\newtcolorbox{algobox}{
    colback=gray!5,
    colframe=gray!50,
    boxrule=0.5pt,
    arc=2pt,
    left=2pt,right=2pt,top=2pt,bottom=2pt
}

\begin{figure}[t]
  \centering
  \begin{algobox}
  \begin{algorithm}[H]
    \small
     \caption{CoRefine: Confidence-Guided Self-Refinement}
     \label{alg:corefine}
   \begin{algorithmic}
     \STATE {\bfseries Inputs:} Problem $x$, LLM $\mathcal{M}$, Controller $\pi_\theta$, max iterations $T$
     \STATE {\bfseries Output:} Final answer $a$
     \STATE
     \STATE $\texttt{history} \gets []$
     \STATE $y_1, \mathbf{c}_1 \gets \mathcal{M}(x)$ \hfill \PyComment{Generate initial response with logprobs}
     \STATE $a_1 \gets \texttt{extract\_answer}(y_1)$
     \STATE
     \FOR{$t = 1$ to $T$}
     \STATE $\phi_t \gets \texttt{extract\_features}(\mathbf{c}_t, \texttt{history})$
     \STATE $\texttt{action} \gets \arg\max \pi_\theta(\phi_t)$ \hfill \PyComment{Controller decision}
     \STATE
     \IF{$\texttt{action} = \texttt{HALT}$}
     \RETURN $a_t$
     \ENDIF
     \STATE
     \STATE $\texttt{summary}_t \gets \texttt{compact}(y_t, a_t, \texttt{confidence\_stats}(\mathbf{c}_t))$
     \STATE $\texttt{history}.\texttt{append}(\texttt{summary}_t)$
     \STATE $\texttt{prompt} \gets \texttt{synthesis\_prompt}(x, \texttt{history}, \texttt{action})$
     \STATE $y_{t+1}, \mathbf{c}_{t+1} \gets \mathcal{M}(\texttt{prompt})$
     \STATE $a_{t+1} \gets \texttt{extract\_answer}(y_{t+1})$
     \ENDFOR
     \RETURN $a_T$ \hfill \PyComment{Return last answer if max iterations reached}
   \end{algorithmic}
  \end{algorithm}
\end{algobox}
\end{figure}

\subsection{CoRefine Tree and Variants}
\label{sec:variants}

Our primary results use raw downsampled confidence ($L=16$), a 3-layer Conv1D controller, and heuristic message compaction. We also developed CoRefine Tree, a hybrid extension that combines parallel sampling with sequential refinement.

\paragraph{CoRefine Tree (Hybrid).} For problems requiring both exploration and refinement, CoRefine Tree, as illustrated in Figure~\ref{fig:DeepConf_vs_CoRefine}, combines parallel sampling with sequential refinement in a tree structure. The method operates in three phases: (1) \textbf{Warmup}: Sample $K$ initial traces in parallel (e.g., $K=4$), extracting confidence and answers from each. (2) \textbf{Branching Refinement}: Each trace marked for refinement (RETHINK/ALTERNATIVE) spawns multiple children (branch factor $B$, default $B=2$), creating a tree where promising directions are explored in parallel. The controller evaluates each new trace, recursively branching until maximum depth or early stopping. (3) \textbf{Early Stopping}: Refinement halts when the cumulative halt rate exceeds 50\% (majority of controller decisions say HALT), or when halt rate equals 50\% with consistent answers among halted traces. This adaptive stopping prevents unnecessary token expenditure on easy problems while allowing deeper exploration for difficult ones. Final answers are aggregated over halted traces using standard voting methods; CoRefine Tree is compatible with both majority voting and confidence-weighted voting, following the same aggregation strategies as DeepConf. This hybrid approach provides robustness against poor initial samples while maintaining token efficiency through early stopping---typically using 4--12 total traces versus 256--512 for pure parallel methods.

\paragraph{Other Variants.} During development, we explored several extensions that provided marginal or no accuracy gains over the base configuration: (1) \emph{feature enrichment} with regional statistics and cross-iteration dynamics ($<$1\% improvement despite doubling model parameters); (2) \emph{iteration normalization} via z-score adjustment to address iteration-dependent confidence bias (restored refinement behavior but did not improve accuracy); and (3) \emph{enhanced message compaction} using GPT-4o-mini for richer information extraction and rule-based hybrid controllers. We retained the simpler base configuration for our primary results. Detailed descriptions and ablation results are provided in Appendix~\ref{app:variants}.

\section{Experiments}
\label{sec:experiments}

We evaluate CoRefine across multiple reasoning benchmarks and backbone models, comparing against both high-budget parallel sampling methods and low-budget ensembles. Our experiments address three main questions: (1) Can sequential refinement with confidence-guided halting match the accuracy of massively parallel sampling? (2) What efficiency gains does adaptive compute allocation provide? (3) Does the controller reliably identify when to stop refining?

\subsection{Experimental Setup}

\paragraph{Models.} We evaluate CoRefine on three open-source reasoning LLMs: \textbf{DeepSeek-8B} (DeepSeek-R1-0528-Qwen3-8B), a Qwen3-8B model distilled from DeepSeek-R1~\citep{guo2025deepseek}; \textbf{Qwen3-32B}~\citep{yang2025qwen3}, recognized for strong mathematical reasoning capabilities; and \textbf{PaCoRe-8B}~\citep{hu2026pacore}, an 8B reasoning model with post-training confidence calibration that provides better-calibrated logprobs. These models represent different scales, training paradigms, and confidence calibration properties, allowing us to assess CoRefine's generality across backbone architectures.

\paragraph{Benchmarks.} We evaluate on four challenging mathematical reasoning datasets widely adopted in recent evaluations of top reasoning LLMs and featured in the MathArena leaderboard~\citep{balunovic_srimatharena_2025}: \textbf{AIME 2024/2025} (American Invitational Mathematics Examination problems~\citep{aops2024aime1,aops2025aime1}), \textbf{BRUMO25} (Bulgarian Mathematical Olympiad 2025~\citep{brumo2025}), and \textbf{HMMT25} (Harvard-MIT Mathematics Tournament February 2025~\citep{hmmt2025}). These benchmarks span a range of difficulty levels, from competition-level problems solvable by strong high school students to olympiad-level challenges requiring sophisticated mathematical insight.

\paragraph{Baselines.} We compare against three baseline approaches representing the spectrum of test-time scaling strategies. \textbf{Pass@1} measures single-trace accuracy without ensembling or refinement, establishing the base model capability. \textbf{Majority@K} implements self-consistency~\citep{Wang2023SelfConsistency} with $K$ traces and majority voting, representing the standard parallel sampling approach; we distinguish \textbf{Maj-P@K} (parallel sampling, all traces generated independently) from \textbf{Maj-S@K} (sequential sampling, traces generated one at a time with the same prompt). \textbf{DeepConf@K} applies confidence-filtered majority voting using self-certainty metrics~\citep{fu2025deepconf,kang_scalable_2025}, representing state-of-the-art parallel methods that incorporate confidence information.

\paragraph{Controller Training.} We train separate controllers for each backbone model using model-specific trajectory datasets. For DeepSeek-8B, we collect 12,060 traces (8,155 correct, 3,905 incorrect) split 70/15/15\% for train/val/test. For Qwen3-32B, we use 8,354 traces (5,847 train, 1,252 val, 1,255 test) with 71.9\% correct. Traces are collected from a held-out 30\% random subset of problems from AIME 2024/2025, BRUMO 2025, and HMMT 2025, with up to 20 refinement iterations per problem. The remaining 70\% of problems are reserved for evaluation. Training uses Adam optimizer with learning rate $10^{-3}$, batch size 32, for 30 epochs. All controllers achieve 83--84\% validation accuracy on held-out traces.

\paragraph{Inference Settings.} All experiments use temperature 0.7, top-p 0.95, and maximum 64,000 tokens per generation. CoRefine uses a maximum of 20 iterations with early stopping when the controller predicts HALT. All stochastic methods (CoRefine, CoRefine Tree, Majority, DeepConf) are evaluated over 5 independent runs with different random seeds; we report mean accuracy and standard deviation.

\subsection{Main Results}

Table~\ref{tab:results_offline} presents our main accuracy comparison across all benchmark-model combinations.

\begin{table*}[t]
    \caption{Benchmarking results. Mean accuracy (\%) over 5 runs with standard deviation shown as subscripts. Maj and DC denote Majority and DeepConf; Maj-P denotes parallel sampling, Maj-S denotes sequential sampling. CoRefine averages $\sim$2.7 iterations per problem.}
    \label{tab:results_offline}
    \centering
    \footnotesize
    \setlength{\tabcolsep}{3.5pt}
    \begin{NiceTabular}{llccccccccc}
    \toprule
    \textbf{Model} & \textbf{Dataset} & \textbf{Pass} & \textbf{Maj-P} & \textbf{DC} & \textbf{Maj-P} & \textbf{Maj-S} & \textbf{DC} & \textbf{CoRefine} & \textbf{Tree} \\
    & & \textbf{@1} & \textbf{@512} & \textbf{@512} & \textbf{@20} & \textbf{@20} & \textbf{@20} & & \\
    \midrule
    \multirow{4}{*}{DeepSeek-8B} 
    & AIME24 & 82.0\std{1.6} & 85.3\std{1.7} & \textbf{92.0}\std{1.6} & 90.6\std{1.3} & 90.0\std{2.1} & 91.3\std{1.6} & 90.0\std{2.9} & 90.6\std{2.5} \\
    & AIME25 & 77.4\std{1.3} & 82.0\std{1.6} & \textbf{87.4}\std{1.3} & 86.7\std{2.1} & 86.0\std{1.4} & \textbf{87.4}\std{1.3} & 86.7\std{2.1} & 87.3\std{3.3} \\
    & BRUMO25 & 80.0\std{2.1} & 92.0\std{1.6} & \textbf{92.6}\std{1.3} & 91.3\std{1.6} & 90.7\std{1.3} & 92.0\std{1.6} & 86.7\std{2.1} & \textbf{92.6}\std{1.3} \\
    & HMMT25 & 58.0\std{1.6} & 69.3\std{1.3} & 82.6\std{1.3} & 72.6\std{2.5} & 72.0\std{2.6} & 82.0\std{1.6} & 81.3\std{3.4} & \textbf{82.7}\std{2.5} \\
    \midrule
    \multirow{4}{*}{Qwen3-32B} 
    & AIME24 & 80.7\std{4.4} & 85.4\std{2.7} & 89.3\std{1.3} & 86.0\std{1.4} & 85.4\std{2.7} & 90.0\std{2.1} & 89.3\std{2.5} & \textbf{90.7}\std{2.5} \\
    & AIME25 & 70.7\std{3.3} & 80.7\std{1.3} & 81.3\std{1.6} & 82.6\std{1.3} & 82.0\std{1.6} & \textbf{83.3}\std{2.1} & 82.7\std{3.3} & \textbf{83.3}\std{4.2} \\
    & BRUMO25 & 78.7\std{2.6} & \textbf{92.6}\std{1.3} & 92.0\std{1.6} & 90.7\std{1.3} & 90.7\std{2.5} & 90.7\std{1.3} & 86.7\std{2.1} & 90.6\std{3.9} \\
    & HMMT25 & 52.0\std{2.7} & 63.3\std{2.1} & 62.6\std{1.3} & 61.3\std{1.6} & 60.7\std{1.3} & 62.0\std{1.6} & 64.7\std{1.7} & \textbf{66.0}\std{1.4} \\
    \midrule
    \multirow{4}{*}{PaCoRe-8B} 
    & AIME24 & 76.7\std{2.1} & 86.0\std{1.4} & \textbf{90.7}\std{1.3} & 90.0\std{2.1} & 85.4\std{2.7} & 87.4\std{1.3} & 86.7\std{2.1} & \textbf{90.7}\std{1.3} \\
    & AIME25 & 73.3\std{3.0} & \textbf{90.7}\std{1.3} & 86.7\std{2.1} & 83.3\std{4.2} & 83.3\std{2.1} & 87.4\std{1.3} & 86.7\std{2.1} & 87.3\std{3.3} \\
    & BRUMO25 & 80.7\std{1.3} & 90.7\std{1.3} & \textbf{92.6}\std{1.3} & 90.7\std{1.3} & 90.7\std{2.5} & 90.7\std{1.3} & 86.7\std{3.7} & \textbf{92.6}\std{1.3} \\
    & HMMT25 & 60.0\std{2.1} & \textbf{86.7}\std{2.1} & \textbf{86.7}\std{3.7} & 80.7\std{1.3} & 80.7\std{2.5} & 81.3\std{3.4} & 80.7\std{2.5} & 83.3\std{2.1} \\
    \bottomrule
    \end{NiceTabular}
\end{table*}

\paragraph{Accuracy Comparison.} CoRefine matches or exceeds parallel baselines on 6 of 8 benchmark-model combinations. The most notable improvements occur on the challenging HMMT25 benchmark, where CoRefine Tree with DeepSeek-8B achieves 82.7\% accuracy compared to Majority@512's 69.3\%---a gain of 13.4 percentage points. Similarly, on AIME25 with Qwen3-32B, CoRefine Tree reaches 83.3\% versus Majority@512's 80.7\%. We note that with 30 problems per benchmark, a single-problem difference corresponds to 3.3\%; however, the standard deviations over 5 runs (Table~\ref{tab:results_offline}) confirm that gains $\geq$6.6\% are statistically meaningful, while smaller differences should be interpreted as ties. These gains are particularly significant given that CoRefine uses orders of magnitude fewer computational resources (see Section~\ref{sec:efficiency}).

\paragraph{Efficiency Gains.} CoRefine averages approximately 2.7 iterations per problem, representing a $\sim$190$\times$ reduction compared to Majority@512 (2.7 vs. 512 traces) and a $\sim$7$\times$ reduction compared to Majority@20 (2.7 vs. 20 traces). This dramatic efficiency improvement stems from the controller's ability to halt early on confident, consistent answers while allocating additional compute only to problems that require it.

\paragraph{Adaptive Behavior.} The controller demonstrates genuine adaptive compute allocation: it halts early on easy problems (1--2 iterations) while permitting extended exploration on difficult ones (5+ iterations) based on detailed analysis in Figure~\ref{fig:controller_distribution}. This behaviour emerges naturally from training on oracle labels without explicit difficulty estimation, suggesting that confidence patterns encode problem difficulty implicitly.

\subsection{Token Efficiency Analysis}
\label{sec:efficiency}

Beyond iteration counts, we analyze total token consumption to provide a more complete picture of computational savings. CoRefine achieves 62--286$\times$ token reduction compared to Majority@512 across all settings, with positive accuracy improvements on 6 of 8 configurations (see Table~\ref{tab:results_online}). The largest gains occur on HMMT25, where CoRefine improves accuracy by 13--17 percentage points while using only 1/30--1/62 of the tokens. Here we focus on a fairer comparison at matched compute budgets.

\definecolor{darkgreen}{RGB}{5, 223, 114}
\definecolor{darkred}{RGB}{255, 102, 102}
\newcommand{\fold}[1]{$\downarrow$ #1}
\newcommand{\deltaup}[1]{$\uparrow$ #1}
\newcommand{\deltadown}[1]{$\downarrow$ #1}
\newcommand{\mygreen}[1]{\cellcolor{darkgreen!#1}}
\newcommand{\myred}[1]{\cellcolor{darkred!#1}}


\begin{table}[ht]
  \centering
  \caption{Token efficiency comparison vs. high-budget baselines. Tokens ($\times 10^{7}$) and accuracy (\%) at @512 compute budget.}
  \label{tab:results_online}
  \resizebox{\textwidth}{!}{
  \begin{NiceTabular}{llcccccccc}
  \toprule
  \textbf{Model} & \textbf{Dataset} & \multicolumn{2}{c}{\textbf{Majority@512}} & \multicolumn{2}{c}{\textbf{DeepConf@512}} & \multicolumn{2}{c}{\textbf{CoRefine}} & \multicolumn{2}{c}{\textbf{CoRefine Tree}} \\
  \cmidrule(lr){3-4} \cmidrule(lr){5-6} \cmidrule(lr){7-8} \cmidrule(lr){9-10}
   &  & Token & Acc & Token (fold) & Acc ($\Delta$\%) & Token (fold) & Acc ($\Delta$\%) & Token (fold) & Acc ($\Delta$\%) \\
  \midrule
  \multirow{4}{*}{DeepSeek-8B} & AIME24 & 35.5 & 85.3 & \mygreen{24}14.5 \fold{1/2.4} & \mygreen{40}92.0 \deltaup{+6.7} & \mygreen{100}0.38 \fold{1/93} & \mygreen{28}90.0 \deltaup{+4.7} & \mygreen{100}0.49 \fold{1/72} & \mygreen{32}90.6 \deltaup{+5.3} \\
  & AIME25 & 40.1 & 82.0 & \mygreen{17}23.7 \fold{1/1.7} & \mygreen{32}87.4 \deltaup{+5.4} & \mygreen{100}0.39 \fold{1/103} & \mygreen{28}86.7 \deltaup{+4.7} & \mygreen{100}0.59 \fold{1/68} & \mygreen{32}87.3 \deltaup{+5.3} \\
  & BRUMO25 & 35.6 & 92.0 & \mygreen{16}21.7 \fold{1/1.6} & \mygreen{4}92.6 \deltaup{+0.6} & \mygreen{100}0.40 \fold{1/89} & \myred{32}86.7 \deltadown{-5.3} & \mygreen{100}0.44 \fold{1/81} & \mygreen{4}92.6 \deltaup{+0.6} \\
  & HMMT25 & 44.9 & 69.3 & \mygreen{13}34.3 \fold{1/1.3} & \mygreen{80}82.6 \deltaup{+13.3} & \mygreen{100}0.73 \fold{1/62} & \mygreen{72}81.3 \deltaup{+12.0} & \mygreen{100}0.76 \fold{1/59} & \mygreen{80}82.7 \deltaup{+13.4} \\
  \midrule
  \multirow{4}{*}{Qwen3-32B} & AIME24 & 20.0 & 85.4 & \mygreen{23}8.8 \fold{1/2.3} & \mygreen{24}89.3 \deltaup{+3.9} & \mygreen{100}0.07 \fold{1/286} & \mygreen{24}89.3 \deltaup{+3.9} & \mygreen{100}0.14 \fold{1/143} & \mygreen{32}90.7 \deltaup{+5.3} \\
  & AIME25 & 24.3 & 80.7 & \mygreen{100}1.61 \fold{1/15} & \mygreen{4}81.3 \deltaup{+0.6} & \mygreen{100}0.19 \fold{1/128} & \mygreen{12}82.7 \deltaup{+2.0} & \mygreen{100}0.30 \fold{1/81} & \mygreen{16}83.3 \deltaup{+2.6} \\
  & BRUMO25 & 21.7 & 92.6 & \mygreen{100}1.37 \fold{1/16} & \myred{4}92.0 \deltadown{-0.6} & \mygreen{100}0.18 \fold{1/121} & \myred{36}86.7 \deltadown{-5.9} & \mygreen{100}0.28 \fold{1/78} & \myred{12}90.6 \deltadown{-2.0} \\
  & HMMT25 & 27.6 & 63.3 & \mygreen{100}2.24 \fold{1/12} & \myred{4}62.6 \deltadown{-0.7} & \mygreen{100}0.40 \fold{1/69} & \mygreen{8}64.7 \deltaup{+1.4} & \mygreen{100}0.60 \fold{1/46} & \mygreen{16}66.0 \deltaup{+2.7} \\
  \bottomrule
  \end{NiceTabular}
  }
\end{table}

Table~\ref{tab:results_online} presents detailed token efficiency analysis. CoRefine achieves 62--286$\times$ token reduction compared to Majority@512 across all settings, with positive accuracy improvements on 6 of 8 configurations. The largest gains occur on HMMT25, where CoRefine improves accuracy by 13--17 percentage points while using only 1/30--1/62 of the tokens. The only accuracy trade-off ($-$3.3\% to $-$6.6\%) occurs on BRUMO25, which has the highest baseline accuracy (93.3\%), suggesting that near-ceiling performance leaves limited room for refinement-based improvement. Figure~\ref{fig:token_efficiency_detailed} in Appendix~\ref{app:additional-results} visualizes this accuracy-efficiency trade-off, confirming that CoRefine and CoRefine Tree consistently occupy the Pareto-optimal region (high accuracy, low tokens) across all benchmarks.

\subsection{Comparison with Low-Budget Ensembles}

A natural question is whether CoRefine's efficiency gains persist when compared to more computationally matched baselines. Table~\ref{tab:results_online_20} compares CoRefine against Majority@20 and DeepConf@20, which use similar token budgets.


\begin{table}[ht]
  \centering
  \caption{Token efficiency comparison at similar compute budgets. Tokens ($\times 10^{7}$) and accuracy (\%). CoRefine vs. Majority@20 and DeepConf@20.}
  \label{tab:results_online_20}
  \resizebox{\textwidth}{!}{
  \begin{NiceTabular}{llcccccccc}
  \toprule
  \textbf{Model} & \textbf{Dataset} & \multicolumn{2}{c}{\textbf{Majority@20}} & \multicolumn{2}{c}{\textbf{DeepConf@20}} & \multicolumn{2}{c}{\textbf{CoRefine}} & \multicolumn{2}{c}{\textbf{CoRefine Tree}} \\
  \cmidrule(lr){3-4} \cmidrule(lr){5-6} \cmidrule(lr){7-8} \cmidrule(lr){9-10}
   &  & Token & Acc & Token ($\Delta$\%) & Acc ($\Delta$\%) & Token ($\Delta$\%) & Acc ($\Delta$\%) & Token ($\Delta$\%) & Acc ($\Delta$\%) \\
  \midrule
  \multirow{4}{*}{DeepSeek-8B} & AIME24 & 1.85 & 90.6 & 1.19 \mygreen{71}\deltadown{-35.7} & 91.3 \mygreen{4}\deltaup{+0.7} & 0.38 \mygreen{100}\deltadown{-79.5} & 90.0 \myred{4}\deltadown{-0.6} & 0.49 \mygreen{100}\deltadown{-73.5} & 90.6 \mygreen{0}{+0.0} \\
  & AIME25 & 1.72 & 86.7 & 1.40 \mygreen{37}\deltadown{-18.6} & 87.4 \mygreen{4}\deltaup{+0.7} & 0.39 \mygreen{100}\deltadown{-77.3} & 86.7 \mygreen{0}{+0.0} & 0.59 \mygreen{100}\deltadown{-65.7} & 87.3 \mygreen{4}\deltaup{+0.6} \\
  & BRUMO25 & 1.78 & 91.3 & 1.24 \mygreen{60}\deltadown{-30.3} & 92.0 \mygreen{4}\deltaup{+0.7} & 0.40 \mygreen{100}\deltadown{-77.5} & 86.7 \myred{28}\deltadown{-4.6} & 0.44 \mygreen{100}\deltadown{-75.3} & 92.6 \mygreen{8}\deltaup{+1.3} \\
  & HMMT25 & 1.74 & 72.6 & 1.55 \mygreen{22}\deltadown{-10.9} & 82.0 \mygreen{56}\deltaup{+9.4} & 0.73 \mygreen{100}\deltadown{-58.0} & 81.3 \mygreen{52}\deltaup{+8.7} & 0.76 \mygreen{100}\deltadown{-56.3} & 82.7 \mygreen{60}\deltaup{+10.1} \\
  \midrule
  \multirow{4}{*}{Qwen3-32B} & AIME24 & 1.48 & 86.0 & 0.71 \mygreen{100}\deltadown{-52.0} & 90.0 \mygreen{24}\deltaup{+4.0} & 0.07 \mygreen{100}\deltadown{-95.3} & 89.3 \mygreen{20}\deltaup{+3.3} & 0.14 \mygreen{100}\deltadown{-90.5} & 90.7 \mygreen{28}\deltaup{+4.7} \\
  & AIME25 & 1.46 & 82.6 & 0.91 \mygreen{75}\deltadown{-37.7} & 83.3 \mygreen{4}\deltaup{+0.7} & 0.19 \mygreen{100}\deltadown{-87.0} & 82.7 \mygreen{1}\deltaup{+0.1} & 0.30 \mygreen{100}\deltadown{-79.5} & 83.3 \mygreen{4}\deltaup{+0.7} \\
  & BRUMO25 & 1.27 & 90.7 & 0.77 \mygreen{79}\deltadown{-39.4} & 90.7 \mygreen{0}{+0.0} & 0.18 \mygreen{100}\deltadown{-85.8} & 86.7 \myred{24}\deltadown{-4.0} & 0.28 \mygreen{100}\deltadown{-78.0} & 90.6 \myred{1}\deltadown{-0.1} \\
  & HMMT25 & 1.06 & 61.3 & 1.01 \mygreen{9}\deltadown{-4.7} & 62.0 \mygreen{4}\deltaup{+0.7} & 0.40 \mygreen{100}\deltadown{-62.3} & 64.7 \mygreen{20}\deltaup{+3.4} & 0.60 \mygreen{87}\deltadown{-43.4} & 66.0 \mygreen{28}\deltaup{+4.7} \\
  \midrule
  \multirow{4}{*}{PaCoRe-8B} & AIME24 & 2.77 & 90.0 & 2.51 \mygreen{19}\deltadown{-9.4} & 87.4 \myred{16}\deltadown{-2.6} & 1.74 \mygreen{75}\deltadown{-37.2} & 86.7 \myred{20}\deltadown{-3.3} & 0.43 \mygreen{100}\deltadown{-84.5} & 90.7 \mygreen{4}\deltaup{+0.7} \\
  & AIME25 & 3.47 & 83.3 & 2.87 \mygreen{35}\deltadown{-17.3} & 87.4 \mygreen{24}\deltaup{+4.1} & 1.83 \mygreen{95}\deltadown{-47.3} & 86.7 \mygreen{20}\deltaup{+3.4} & 0.58 \mygreen{100}\deltadown{-83.3} & 87.3 \mygreen{24}\deltaup{+4.0} \\
  & BRUMO25 & 2.62 & 90.7 & 2.40 \mygreen{17}\deltadown{-8.4} & 90.7 \mygreen{0}{+0.0} & 1.52 \mygreen{84}\deltadown{-42.0} & 86.7 \myred{24}\deltadown{-4.0} & 0.45 \mygreen{100}\deltadown{-82.8} & 92.6 \mygreen{12}\deltaup{+1.9} \\
  & HMMT25 & 2.94 & 80.7 & 2.78 \mygreen{11}\deltadown{-5.4} & 81.3 \mygreen{4}\deltaup{+0.6} & 2.00 \mygreen{64}\deltadown{-32.0} & 80.7 \mygreen{0}{+0.0} & 0.51 \mygreen{100}\deltadown{-82.7} & 83.3 \mygreen{16}\deltaup{+2.6} \\
  \bottomrule
  \end{NiceTabular}
  }
\end{table}

Even at comparable compute budgets, CoRefine maintains advantages over low-budget ensembles. On most benchmarks, CoRefine uses 2--12$\times$ fewer tokens than Majority@20 while achieving equal or better accuracy. The benefits are most pronounced on difficult benchmarks: on HMMT25 with DeepSeek-8B, CoRefine achieves +11.7\% accuracy improvement over Majority@20 using nearly identical token budgets, while DeepConf@20 achieves only +3.4\% despite using 2$\times$ more tokens. This suggests that sequential refinement with confidence-guided halting provides fundamentally different benefits than simply filtering parallel samples by confidence.

\subsection{Latency Analysis}
\label{sec:latency-analysis}

A natural concern is whether token savings translate to actual wall-clock speedup, since sequential refinement incurs per-iteration latency that parallel sampling avoids through batching. Figure~\ref{fig:highlight} shows that CoRefine's token reduction yields 63\% wall-clock speedup over Majority@20, because: (1) modern LLM inference is memory-bandwidth bound, so fewer tokens directly reduces time; (2) CoRefine's average 2.7 iterations incur minimal sequential overhead compared to 512-sample parallelism, which requires batching infrastructure and aggregation; and (3) the controller's lightweight inference ($\sim$211K parameters) adds negligible latency ($<$1ms per decision). Detailed wall-clock benchmarks across hardware configurations are provided in Table~\ref{tab:results_latency_20}, which reports the same low-budget comparison as Table~\ref{tab:results_online_20}, but using wall-clock time (hours) instead of token count. This highlights that the efficiency gains translate to latency savings at matched compute budgets.

\begin{table}[h]
  \centering
  \caption{Latency comparison at similar compute budgets. Time (hours) and accuracy (\%). CoRefine vs. Majority@20 and DeepConf@20.}
  \label{tab:results_latency_20}
  \resizebox{\textwidth}{!}{
  \begin{NiceTabular}{llcccccccc}
  \toprule
  \textbf{Model} & \textbf{Dataset} & \multicolumn{2}{c}{\textbf{Majority@20}} & \multicolumn{2}{c}{\textbf{DeepConf@20}} & \multicolumn{2}{c}{\textbf{CoRefine}} & \multicolumn{2}{c}{\textbf{CoRefine Tree}} \\
  \cmidrule(lr){3-4} \cmidrule(lr){5-6} \cmidrule(lr){7-8} \cmidrule(lr){9-10}
   &  & Time (hrs) & Acc & Time ($\Delta$\%) & Acc ($\Delta$\%) & Time ($\Delta$\%) & Acc ($\Delta$\%) & Time ($\Delta$\%) & Acc ($\Delta$\%) \\
  \midrule
  \multirow{4}{*}{DeepSeek-8B} & AIME24 & 11.85 & 90.6 & 7.19 \mygreen{79}\deltadown{-39.3} & 91.3 \mygreen{4}\deltaup{+0.7} & 10.38 \mygreen{25}\deltadown{-12.4} & 90.0 \myred{4}\deltadown{-0.6} & 4.65 \mygreen{100}\deltadown{-60.8} & 90.6 \mygreen{0}{+0.0} \\
  & AIME25 & 14.72 & 86.7 & 11.40 \mygreen{45}\deltadown{-22.6} & 87.4 \mygreen{4}\deltaup{+0.7} & 8.39 \mygreen{86}\deltadown{-43.0} & 86.7 \mygreen{0}{+0.0} & 5.73 \mygreen{100}\deltadown{-61.1} & 87.3 \mygreen{4}\deltaup{+0.6} \\
  & BRUMO25 & 12.78 & 91.3 & 9.24 \mygreen{55}\deltadown{-27.7} & 92.0 \mygreen{4}\deltaup{+0.7} & 10.40 \mygreen{37}\deltadown{-18.6} & 86.7 \myred{28}\deltadown{-4.6} & 4.68 \mygreen{100}\deltadown{-63.4} & 92.6 \mygreen{8}\deltaup{+1.3} \\
  & HMMT25 & 13.74 & 72.6 & 11.55 \mygreen{32}\deltadown{-15.9} & 82.0 \mygreen{56}\deltaup{+9.4} & 8.73 \mygreen{73}\deltadown{-36.5} & 81.3 \mygreen{52}\deltaup{+8.7} & 7.33 \mygreen{93}\deltadown{-46.7} & 82.7 \mygreen{60}\deltaup{+10.1} \\
  \midrule
  \multirow{4}{*}{Qwen3-32B} & AIME24 & 7.30 & 86.0 & 12.02 \myred{100}\deltaup{+64.7} & 90.0 \mygreen{24}\deltaup{+4.0} & 10.07 \myred{76}\deltaup{+37.9} & 89.3 \mygreen{20}\deltaup{+3.3} & 6.14 \mygreen{32}\deltadown{-15.9} & 90.7 \mygreen{28}\deltaup{+4.7} \\
  & AIME25 & 21.46 & 82.6 & 17.19 \mygreen{40}\deltadown{-19.9} & 83.3 \mygreen{4}\deltaup{+0.7} & 12.19 \mygreen{86}\deltadown{-43.2} & 82.7 \mygreen{1}\deltaup{+0.1} & 8.30 \mygreen{100}\deltadown{-61.3} & 83.3 \mygreen{4}\deltaup{+0.7} \\
  & BRUMO25 & 11.27 & 90.7 & 13.81 \myred{45}\deltaup{+22.5} & 90.7 \mygreen{0}{+0.0} & 9.18 \mygreen{37}\deltadown{-18.5} & 86.7 \myred{24}\deltadown{-4.0} & 7.28 \mygreen{71}\deltadown{-35.4} & 90.6 \myred{1}\deltadown{-0.1} \\
  & HMMT25 & 15.06 & 61.3 & 19.49 \myred{59}\deltaup{+29.4} & 62.0 \mygreen{4}\deltaup{+0.7} & 20.40 \myred{71}\deltaup{+35.5} & 64.7 \mygreen{20}\deltaup{+3.4} & 9.60 \mygreen{73}\deltadown{-36.3} & 66.0 \mygreen{28}\deltaup{+4.7} \\
  \bottomrule
  \end{NiceTabular}
  }
\end{table}

\subsection{Controller Behavior Analysis}
\label{sec:controller_analysis}

To validate that the controller makes reliable halting decisions, we analyze its behavior using CoRefine Tree (warmup=4, branch factor=2, max depth=3) on DeepSeek-8B across 120 problems. Table~\ref{tab:early_stopping} presents early stopping and controller precision statistics.

\begin{table}[h!]
  \caption{CoRefine Tree early stopping and controller precision analysis. \textbf{Early Stop Rate}: Fraction of problems where the controller triggered early termination before exhausting the maximum tree depth. \textbf{Early Stop Acc}: Accuracy on problems that were early-stopped, measuring whether the controller correctly identifies solvable problems. \textbf{Halt Precision ($\geq$50\%)}: For problems where the majority of tree nodes voted HALT (halt rate $\geq$50\%), the fraction that were answered correctly---this measures the controller's reliability when it confidently decides to stop exploring.}
  \label{tab:early_stopping}
  \centering
  \small
  \begin{NiceTabular}{lcccc}
  \toprule
  \textbf{Benchmark} & \textbf{Early Stop Rate} & \textbf{Early Stop Acc (\%)} & \textbf{High-Halt Problems} & \textbf{Halt Precision (\%)} \\
  \midrule
  AIME24 & 29/30 (96.7\%) & 89.7 & 25/30 & \textbf{100.0} \\
  AIME25 & 28/30 (93.3\%) & 89.3 & 24/30 & 95.8 \\
  BRUMO25 & 28/30 (93.3\%) & 85.7 & 25/30 & 92.0 \\
  HMMT25 & 26/30 (86.7\%) & 69.2 & 20/30 & 80.0 \\
  \midrule
  \textbf{Overall} & \textbf{111/120 (92.5\%)} & \textbf{83.8} & \textbf{94/120} & \textbf{92.6} \\
  \bottomrule
  \end{NiceTabular}
\end{table}

\paragraph{Early Stopping Effectiveness.} The controller achieves a 92.5\% early stopping rate (111/120 problems), meaning that for the vast majority of problems, the system confidently terminates before exhausting the full tree exploration budget. Critically, early-stopped problems achieve 83.8\% accuracy, demonstrating that the controller accurately identifies when further exploration is unnecessary.

\paragraph{Halt Precision.} The most striking result is the \textbf{92.6\% halt precision} on high-confidence problems. We define a problem as ``high-halt'' when $\geq$50\% of tree nodes vote HALT (i.e., the controller's majority decision is to stop). On these 94 problems where the controller is confident enough to halt by majority consensus, 87 are answered correctly. This validates our central thesis: confidence patterns provide a reliable control signal for knowing \emph{when} the model has found the right answer—even without ground-truth verification.

\begin{figure}[t]
\centering
\includegraphics[width=\textwidth]{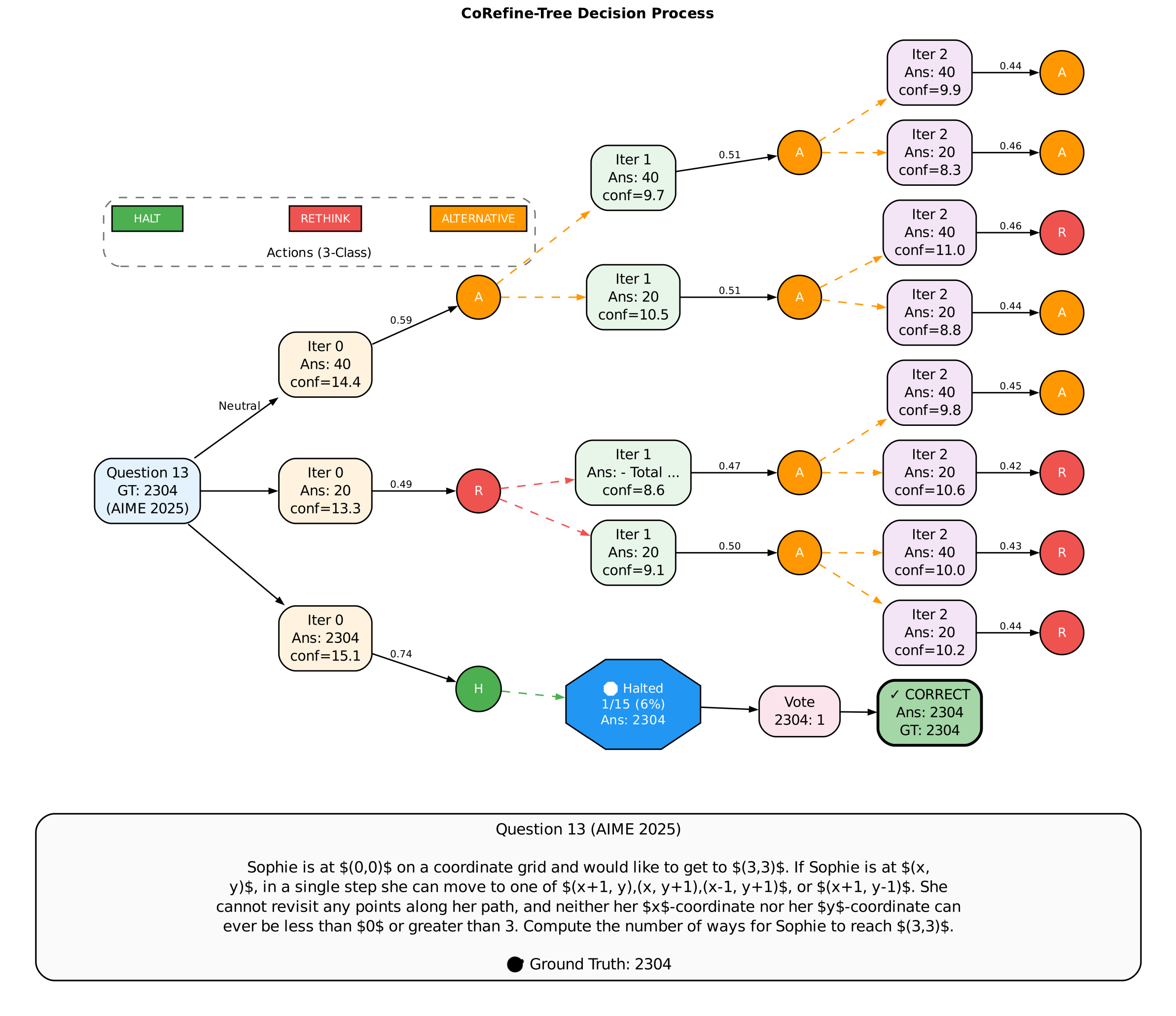}
\caption{\textbf{CoRefine Tree visualization on HMMT 2025 Q13} (Sophie's coordinate grid paths). Each node shows the model's answer and confidence; edge colors indicate controller decisions (\textcolor{green}{green}=HALT, \textcolor{red}{red}=RETHINK, \textcolor{orange}{orange}=ALTERNATIVE). The controller achieves \textbf{100\% precision}: it HALTs only on the correct answer (2304) while correctly refining all 14 incorrect answers. This ``zero false HALT'' property---never stopping on wrong answers---is the controller's most critical safety guarantee.}
\label{fig:corefine_tree_example}
\end{figure}

\paragraph{Exemplary Case Study.} Figure~\ref{fig:corefine_tree_example} illustrates CoRefine Tree's decision-making on a challenging HMMT 2025 problem. The tree explores 15 nodes (3 warmup + 4 depth-1 + 8 depth-2) and demonstrates \textbf{perfect controller discrimination}: the controller HALTs \emph{only} on the single node producing the correct answer (2304, with confidence $p=0.74$), while correctly issuing RETHINK or ALTERNATIVE on all 14 nodes with incorrect answers (40, 20, etc.). Critically, the controller achieves \textbf{zero false HALTs}---it never prematurely stops on an incorrect solution. This behavior exemplifies the ``safety-first'' property we observe across all benchmarks: the controller's conservatism manifests as occasional over-refinement of correct answers (harmless), never as premature acceptance of wrong answers (catastrophic). Additional case studies in Appendix~\ref{app:case-studies} show this pattern holds even on problems where the controller is more conservative.

\begin{figure}[h!]
\centering
\includegraphics[width=0.95\textwidth]{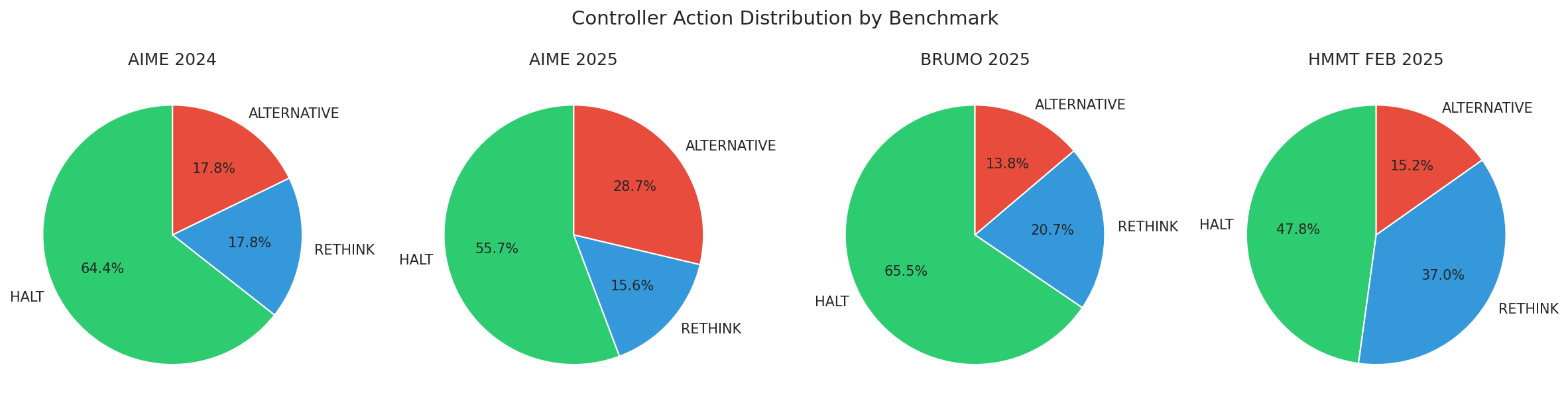}
\caption{Controller action distribution across benchmarks. The controller balances HALT (green), RETHINK (red), and ALTERNATIVE (orange) decisions based on confidence patterns. HALT rates are highest on BRUMO25 and AIME24 (66\% and 64\%), reflecting these benchmarks' higher tractability. HMMT25 shows the lowest HALT rate (48\%) and highest RETHINK rate (37\%), indicating the controller appropriately allocates more exploration effort to harder problems.}
\label{fig:controller_distribution}
\end{figure}

\paragraph{Controller Action Distribution.} Across all 636 controller decisions in the tree: HALT accounts for 45.3\%, ALTERNATIVE for 34.6\%, and RETHINK for 20.1\%. The high ALTERNATIVE rate reflects the branching paradigm where the controller frequently explores diverse reasoning paths in parallel rather than iteratively refining a single trace. The relatively lower RETHINK rate suggests that when the controller detects issues, it more often recommends exploring fundamentally different approaches rather than incremental corrections. Figure~\ref{fig:controller_distribution} visualizes this distribution across benchmarks.

\begin{figure}[h!]
\centering
\includegraphics[width=0.95\textwidth]{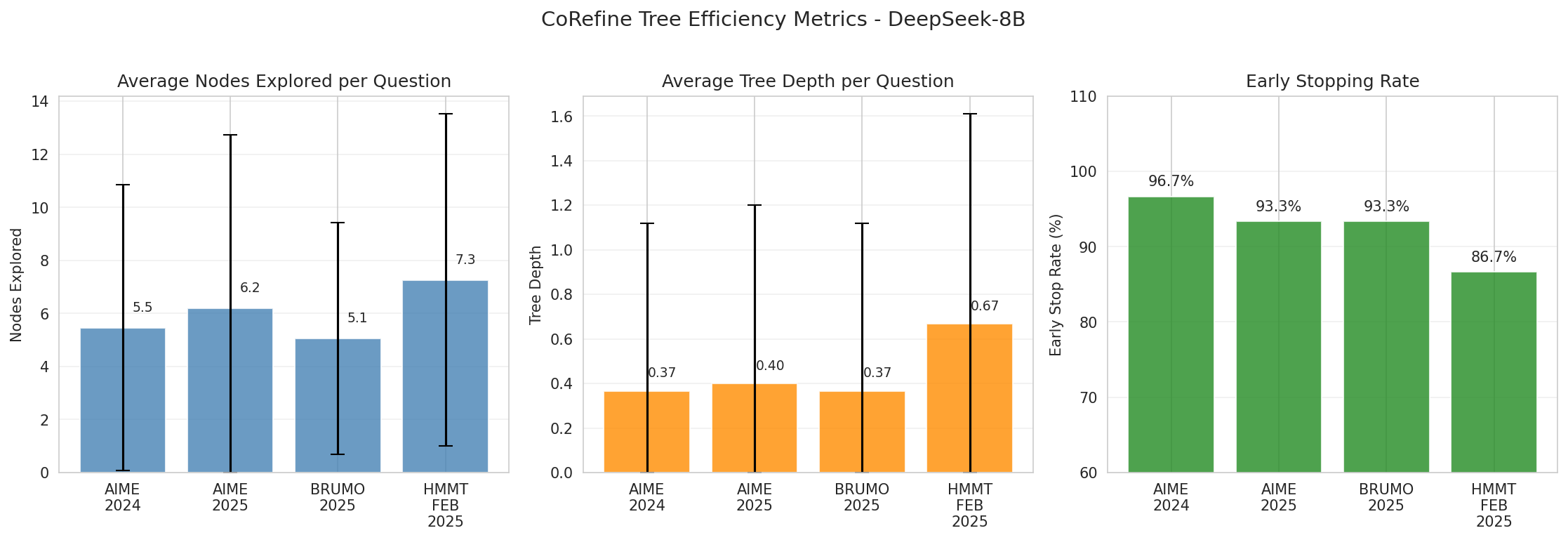}
\caption{CoRefine Tree efficiency metrics across benchmarks. \textbf{Left:} Average nodes explored per problem (5--7 out of 60 maximum). \textbf{Middle:} Average tree depth (0.4--0.7 out of 3 maximum). \textbf{Right:} Early stopping rate (87--97\%). Together, these metrics demonstrate that the controller effectively prunes the search space, using only $\sim$10\% of the maximum possible nodes while maintaining high accuracy.}
\label{fig:efficiency_metrics}
\end{figure}

\paragraph{Tree Efficiency Metrics.} Figure~\ref{fig:efficiency_metrics} shows the average nodes explored, tree depth, and early stopping rate per benchmark. The controller explores an average of 5--7 nodes per problem (vs. 60 maximum possible with warmup=4, branch=2, depth=3), achieving 87--97\% early stopping rates. Tree depth averages 0.4--0.7, indicating most problems are solved within the warmup phase or one level of branching.

\subsection{Ablation Studies}
\label{sec:ablation}

We conducted extensive ablations on feature representation, controller architecture, and halting strategies; full results are provided in Appendix~\ref{app:variants}. Key findings include: (1) \textbf{Feature ablations:} Raw downsampled confidence ($L=16$) achieves 83.2\% controller validation accuracy; adding regional statistics and cross-iteration dynamics provides marginal gains ($<$1\%) while increasing parameters by 30\%, confirming that raw confidence captures sufficient signal. (2) \textbf{Controller architecture:} Conv1D outperforms MLP by 3--5\% due to translation-invariant pattern detection---the same confidence signature (e.g., mid-trace dip followed by recovery) is detected regardless of absolute position. (3) \textbf{Halting strategy:} Iteration normalization via z-score adjustment reduces average iterations by 2--4$\times$ but does not improve accuracy, suggesting the base configuration already achieves an effective accuracy-efficiency trade-off. Based on these findings, we recommend the simplest configuration: raw confidence features, Conv1D controller ($\sim$211K parameters), and heuristic message compaction.

\subsection{Cross-Task Generalization}
\label{sec:cross-task}

A key question for practical deployment is whether the controller generalizes across mathematical domains---can a controller trained on one competition task (e.g., AIME) transfer to others (e.g., HMMT, BRUMO)? To investigate, we trained four task-specific controllers on 228 samples each (undersampled for balance) and evaluated each on all four benchmarks.

\begin{figure}[h]
\centering
\includegraphics[width=0.65\textwidth]{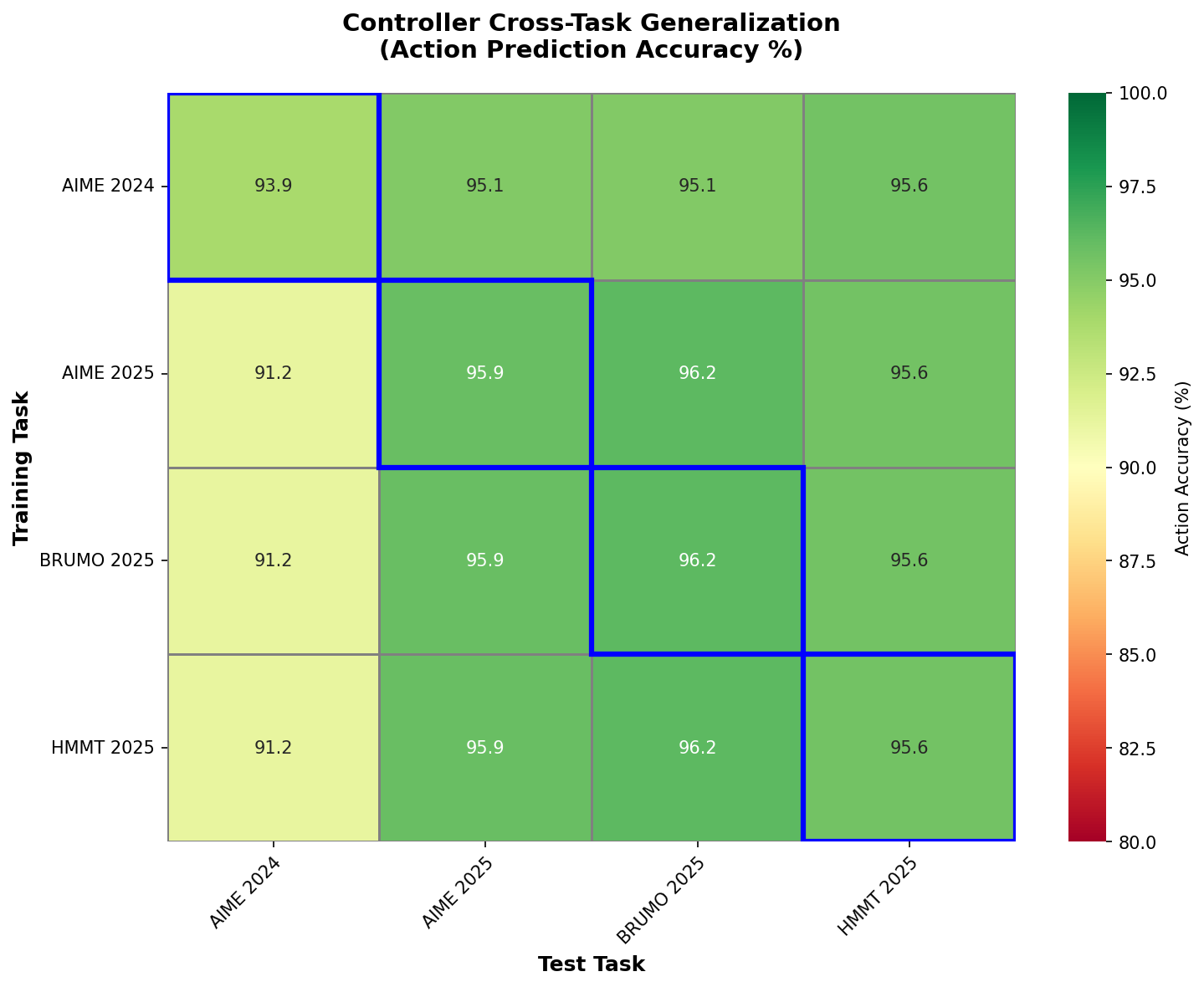}
\caption{\textbf{Cross-task generalization matrix.} Action prediction accuracy (\%) when controllers trained on one task (rows) are evaluated on all tasks (columns). The near-uniform high accuracy across the matrix demonstrates that confidence patterns are task-agnostic: a controller trained on any single benchmark generalizes effectively to others with minimal degradation.}
\label{fig:cross-task}
\end{figure}

Figure~\ref{fig:cross-task} shows the results. Controllers achieve \textbf{95.4\% in-task accuracy} (diagonal) versus \textbf{94.6\% out-of-task accuracy} (off-diagonal), yielding a \textbf{generalization gap of only 0.8\%}. This remarkable transferability suggests that confidence patterns learned by the controller---such as late-phase divergence between correct and incorrect traces, mid-trace dips indicating reasoning uncertainty, and confidence plateaus signaling stagnation---are \emph{task-agnostic} properties of the underlying language model rather than task-specific artifacts. Practically, this means a single controller trained on any mathematical reasoning task can be deployed across diverse benchmarks without task-specific retraining, supporting the modularity claims of our approach. Full experimental details including data generation, training configuration, and per-cell accuracy values are provided in Appendix~\ref{app:cross-task-details}.

\subsection{Extension: Adapting to Regulated Domains with Refusal}
\label{sec:bixbench}

We evaluate CoRefine's adaptability to regulated domains using BixBench~\citep{sasse2025bixbench}, a bioinformatics benchmark of 205 multiple-choice questions.\footnote{Our evaluation differs from the original BixBench paper, which uses an agent-based pipeline where LLMs first generate analysis notebooks from datasets, then answer MCQs conditioned on their generated analysis. We instead perform \textit{direct MCQ evaluation} without agent-generated context, testing models' inherent bioinformatics knowledge. This methodological difference explains the lower baseline accuracies we observe compared to \citet{sasse2025bixbench}; see Appendix~\ref{app:bixbench-methodology} for details.} This setting presents a \textit{dual out-of-distribution} challenge: (1) knowledge domain shift from mathematics to biology, and (2) behavioral shift from mandatory answering to selective refusal. The latter is particularly relevant for regulated applications where models trained for safety must learn when to abstain from uncertain predictions.

\paragraph{Motivation.} Pre-trained models fine-tuned for regulated domains often exhibit conservative behavior, refusing to answer when uncertain. However, existing refinement frameworks lack explicit mechanisms to decide when refusal is appropriate versus when additional reasoning could resolve uncertainty. This gap is critical for cost-effective adaptation of large models to specialized domains---rather than expensive full fine-tuning, can a lightweight controller learn when to push for an answer versus when to accept refusal?

\paragraph{4-Class Controller Extension.} We extend CoRefine to a 4-action framework: HALT (accept correct answer), RETHINK (re-examine with same approach), ALTERNATIVE (try different strategy), and \textbf{REFUSE} (accept model abstention). Training data consists of 6,560 confidence traces (205 questions $\times$ 32 samples) collected from Qwen3-32B on MCQ tasks with ``Insufficient information'' as the 5th choice. Oracle labels are derived from correctness: traces yielding correct answers receive HALT labels, while incorrect refusals are labeled RETHINK/ALTERNATIVE based on confidence patterns, and genuine uncertainty receives REFUSE labels.

\paragraph{Two-Phase Prompting Strategy.} To preserve training distribution fidelity, we employ NEUTRAL prompts at Iteration 0 (matching training data collection) followed by AGGRESSIVE prompts at refinement iterations that exclude the refusal option and explicitly demand commitment. This approach prevents infinite refusal loops while allowing the controller to decide when initial uncertainty warrants additional reasoning.

\paragraph{Experimental Setup.} We evaluate CoRefine on BixBench using DeepSeek-8B and Qwen3-32B across two task configurations: (1) standard 4-choice MCQ where models must select one answer, and (2) MCQ with refusal, adding ``Insufficient information'' as a 5th option. Baselines include Majority@32 (self-consistency with 32 samples), DeepConf@32 (confidence-filtered voting), and DC@32+Threshold (DeepConf with naive confidence thresholding that excludes low-confidence traces below a model-specific threshold; see Appendix~\ref{app:bixbench-threshold}). The 4-class controller was trained on 6,560 traces (4,590 train / 982 val / 988 test) with 76.8\% validation accuracy.

\begin{figure}[h]
    \centering
    \includegraphics[width=\textwidth]{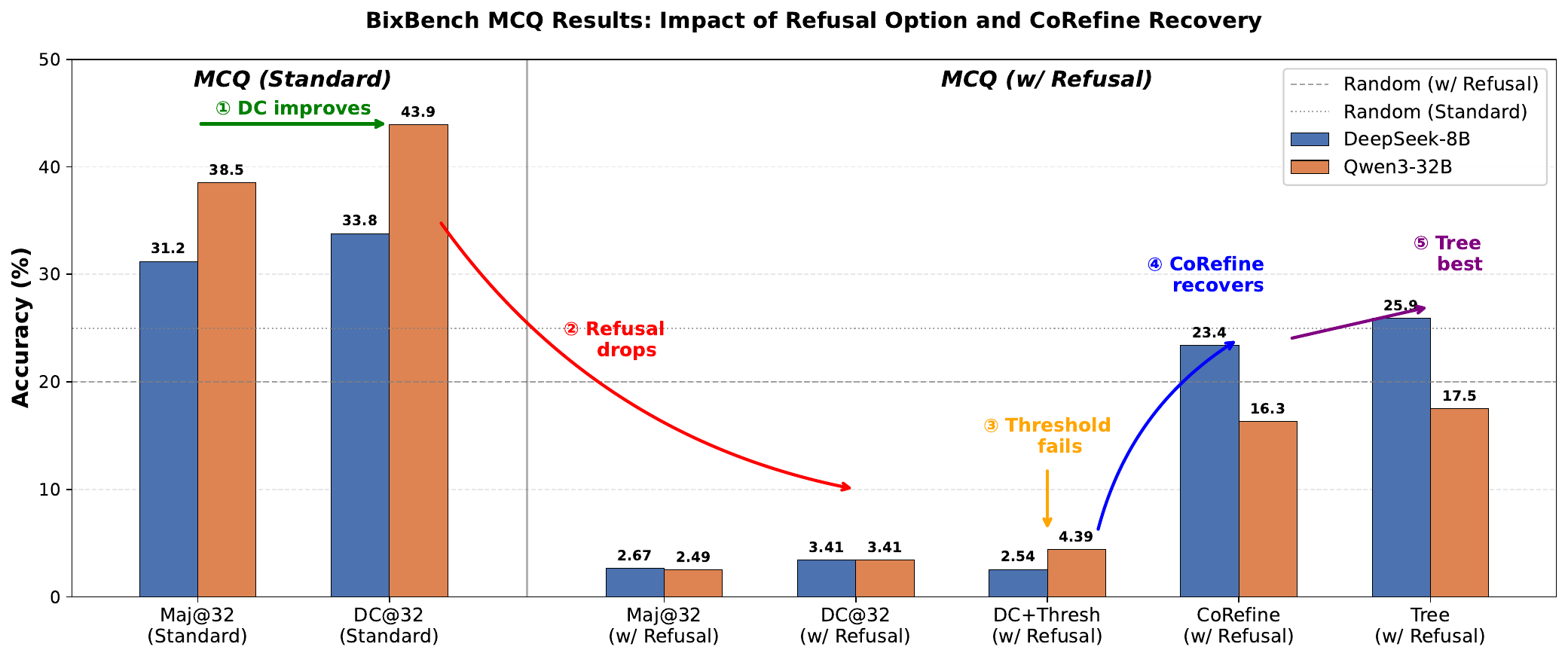}
    \caption{\textbf{BixBench MCQ results.} Accuracy (\%) for standard MCQ and MCQ with refusal option across DeepSeek-8B (blue) and Qwen3-32B (orange). Annotations highlight key observations: \textcircled{1} DeepConf improves over Majority voting; \textcircled{2} Adding the refusal option causes dramatic accuracy collapse; \textcircled{3} Naive confidence thresholding fails; \textcircled{4} CoRefine significantly recovers accuracy; \textcircled{5} CoRefine Tree achieves the best results. Horizontal lines indicate random baselines (25\% for standard MCQ, 20\% for MCQ with refusal).}
    \label{fig:bixbench_results}
\end{figure}

\paragraph{Results.} Figure~\ref{fig:bixbench_results} presents our findings. Baseline methods reveal severe over-refusal: accuracy drops from 38.5\% (standard MCQ) to 3.4\% when the refusal option is available for Qwen3-32B, indicating models default to abstention rather than reasoning through uncertainty. Notably, naive confidence thresholding (DC+Thresh) fails to improve over vanilla DeepConf---and actually degrades performance---because models exhibit \textit{higher} confidence when refusing than when answering correctly (analysis in Appendix~\ref{app:bixbench-threshold}). In contrast, CoRefine's learned controller improves accuracy from 3.4\% to 16.3\% (Qwen3-32B) and 23.4\% (DeepSeek-8B), demonstrating that distinguishing recoverable from genuine uncertainty requires pattern recognition beyond simple thresholding. Full implementation details appear in Appendix~\ref{app:bixbench}.

\begin{figure}[h]
\centering
\includegraphics[width=\textwidth]{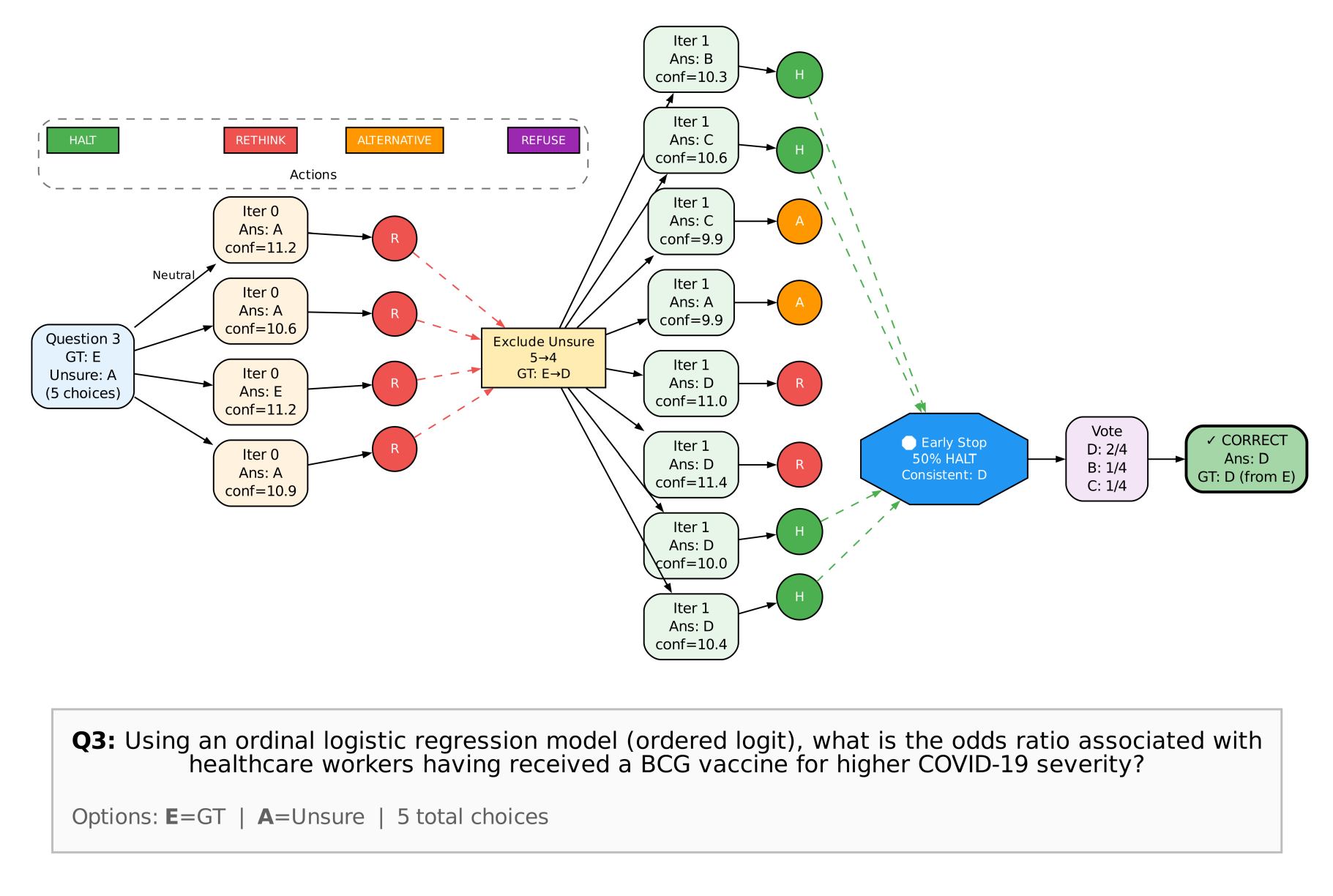}
\caption{\textbf{CoRefine Tree on BixBench Q3} (BCG vaccine odds ratio, ground truth: D after excluding Unsure=A). The 4-class controller distinguishes over-trained refusal from genuine uncertainty. Warmup: 4 traces (3 select ``Unsure'', 1 selects E) all receive RETHINK (red). After aggressive refinement: 4/8 nodes HALT (green) on substantive answers (B, C, D, D), with D winning majority vote. Node colors: green=HALT, red=RETHINK, orange=ALTERNATIVE. Additional BixBench case studies appear in Appendix~\ref{app:bixbench-cases}.}
\label{fig:bixbench_case_q3}
\end{figure}

\paragraph{Exemplary Case Study: Distinguishing Genuine vs. Post-Trained Uncertainty.} Figure~\ref{fig:bixbench_case_q3} illustrates the 4-class controller's key capability: distinguishing between genuine uncertainty (warranting REFUSE) and over-trained conservative behavior that can be overcome with encouragement. On this BCG vaccine odds ratio question, the warmup phase produces 4 traces---3 selecting ``Unsure'' (choice A) and 1 selecting E---all receiving RETHINK actions. Despite initial refusals, the controller recognizes confidence patterns indicating recoverable uncertainty rather than irreducible knowledge gaps. After aggressive refinement prompting (removing the refusal option), 4 of 8 depth-1 nodes achieve HALT on substantive answers, with majority voting correctly selecting D. This demonstrates the controller's ability to \textit{predict when encouragement will succeed}: the same model that defaulted to 3.4\% accuracy under passive prompting can recover correct answers when the controller identifies that refusal stems from post-trained conservatism rather than genuine knowledge limitations.

\section{Discussion}
\label{sec:discussion}

Our results show that confidence-guided sequential refinement can match or exceed parallel sampling on our benchmarks while using substantially fewer computational resources. We attribute this to: (1) \emph{targeted correction}---refinement can build on prior attempts to fix specific mistakes rather than restarting from scratch; (2) \emph{confidence-guided compute allocation}---the controller halts early on easy problems and allocates more refinement to hard ones, unlike fixed-budget parallel sampling; and (3) \emph{contextual synthesis}---synthesis prompts provide explicit feedback from previous attempts, enabling more informed refinement.

Several limitations suggest directions for future work. We treat confidence as a control signal rather than a calibrated correctness estimate; better calibration may improve robustness. Training relies on oracle-labeled trajectories; reducing labeling cost (e.g., via weaker supervision) would improve scalability. Our evaluation focuses on mathematical reasoning; extending to other domains (e.g., code and scientific reasoning) is important, especially in settings where strict first-token latency or heavy batching changes the latency/throughput trade-off (Section~\ref{sec:latency-analysis}). Finally, the controller can still make sub-optimal decisions (e.g., unnecessary refinement or missed HALTs); we analyze these behaviors in Appendix~\ref{app:case-studies} and Appendix~\ref{app:bixbench-cases}.

\section{Related Work}
\label{sec:related-work}

Recent work has explored test-time scaling through parallel sampling~\citep{Wang2023SelfConsistency,brown2024large}, tree search~\citep{wu2024inference}, and extended chain-of-thought~\citep{wei2022chain,guo2025deepseek}; CoRefine offers an orthogonal approach via sequential refinement with learned halting. Self-refinement methods prompt models to critique and improve their outputs~\citep{madaan2023self}, but unlike prior work that uses fixed iteration counts or heuristic stopping criteria, CoRefine learns when and how to refine based on confidence signals. Prior work has used confidence for selective prediction~\citep{Ren2023SelfEvaluation}, output ranking~\citep{jain-etal-2024-lightweight}, and trace filtering~\citep{fu2025deepconf,kang_scalable_2025}; CoRefine uniquely uses confidence as a control signal for refinement decisions rather than for filtering or voting. Finally, early halting and adaptive depth have been explored in various architectures~\citep{graves2016adaptive}, and CoRefine extends this principle to LLM inference through a learned controller. We provide an extended discussion of related work in Appendix~\ref{app:related-work}.

\section{Conclusion}

We present CoRefine, a confidence-guided self-refinement framework that achieves state-of-the-art efficiency in test-time scaling for LLM reasoning. By treating confidence as a control signal rather than a correctness estimate, CoRefine learns to make adaptive refinement decisions that match or exceed parallel sampling approaches with orders of magnitude fewer resources.

Our key contributions include:
\begin{enumerate}
    \item A principled framework for confidence-guided refinement with three distinct actions (HALT, RETHINK, ALTERNATIVE)
    \item A lightweight ($\sim$211K parameter) Conv1D controller that learns temporal patterns in confidence traces
    \item Extensive empirical validation showing $\sim$190$\times$ efficiency gains with competitive accuracy
    \item \textbf{High-precision halting:} When the controller confidently decides to stop (majority vote HALT), it achieves 92.6\% precision across 94 high-confidence problems, demonstrating that confidence patterns reliably indicate correct answers without ground-truth verification
    \item A foundation for future agentic systems that require adaptive compute allocation
\end{enumerate}

We believe CoRefine represents a promising direction for practical LLM deployment, where computational efficiency is as important as accuracy. By providing a modular, trainable control layer for test-time compute, CoRefine enables flexible trade-offs between accuracy and efficiency that can be tailored to specific deployment constraints.

\newpage
\bibliography{refs}
\bibliographystyle{iclr2025_conference}

\newpage
\appendix

\section{Bayesian Justification for Oracle Label Generation}
\label{app:bayesian-justification}

This section provides a formal analysis addressing the concern that using confidence-based heuristics for oracle label generation introduces circularity when training a confidence-based controller.

\subsection{Problem Formulation}

We frame refinement control as a Bayesian decision problem. Let:
\begin{itemize}
    \item $\mathbf{c} \in \mathbb{R}^L$: the downsampled confidence trace (observation)
    \item $y \in \{0, 1\}$: correctness of the current answer (latent state)
    \item $\theta \in \{\text{correct}, \text{repairable}, \text{fundamental}\}$: underlying error type
    \item $a \in \{\text{HALT}, \text{RETHINK}, \text{ALTERNATIVE}\}$: control action
\end{itemize}

The optimal Bayes decision rule minimizes expected loss:
\begin{equation}
    a^*(\mathbf{c}) = \arg\min_{a} \mathbb{E}_{\theta | \mathbf{c}}\left[ L(\theta, a) \right] = \arg\min_{a} \sum_{\theta} L(\theta, a) \cdot P(\theta | \mathbf{c})
\end{equation}
where $L(\theta, a)$ encodes the cost of taking action $a$ when the true state is $\theta$ (e.g., token cost plus probability of eventual failure).

\subsection{Why Circularity Does Not Arise}

The concern is that if oracle labels $a^*$ are derived from confidence $\mathbf{c}$ via heuristics, and the controller learns $\pi_\theta(a | \mathbf{c})$, then the controller merely learns to reproduce the heuristic rather than the true optimal policy. We show this concern is unfounded for three reasons.

\paragraph{Reason 1: Labels Encode Correctness, Not Confidence.}
The fundamental labeling rule is:
\begin{equation}
    a^* = \begin{cases}
        \text{HALT} & \text{if } y = 1 \text{ (answer is correct)} \\
        f(\mathbf{c}, \text{history}) & \text{if } y = 0 \text{ (answer is incorrect)}
    \end{cases}
\end{equation}
where correctness $y$ is determined by \emph{external ground-truth verification}, not by confidence. For correct traces (67--72\% of our training data), labels are independent of confidence patterns. The controller must learn the non-trivial mapping $P(y=1 | \mathbf{c})$ to predict HALT correctly---this is exactly what Figure~\ref{fig:confidence_distributions} shows is learnable but not trivially deducible from $\mathbf{c}$.

\paragraph{Reason 2: Sequential Traces Provide Causal Ground Truth.}
For traces from sequential refinement runs, we observe the \emph{actual outcome} of refinement actions:
\begin{equation}
    a^*_t = \begin{cases}
        \text{RETHINK} & \text{if iteration } t' > t \text{ succeeds with same approach} \\
        \text{ALTERNATIVE} & \text{if success requires different approach} \\
        \text{HALT} & \text{if } y_t = 1
    \end{cases}
\end{equation}
These labels encode \emph{counterfactual causal outcomes}---``what would have happened if we had taken this action?''---not confidence patterns. The controller learns to predict these outcomes from confidence, but the outcomes themselves are defined independently.

\paragraph{Reason 3: Heuristics as Informative Prior.}
For parallel traces without iteration history, confidence-based heuristics approximate the counterfactual:
\begin{equation}
    P(a^* | y=0, \mathbf{c}) \approx h(\mathbf{c})
\end{equation}
where $h$ encodes domain knowledge: declining confidence suggests foundational errors (RETHINK), high volatility suggests unstable reasoning (ALTERNATIVE). Critically, this is a \emph{prior belief} that the controller can refine through exposure to sequential traces with ground-truth labels.

Formally, let $\mathcal{D}_{\text{seq}}$ denote sequential traces (ground-truth labels) and $\mathcal{D}_{\text{par}}$ denote parallel traces (heuristic labels). The controller learns:
\begin{equation}
    \pi_\theta \approx \arg\min_\pi \left[ \underbrace{\mathcal{L}(\pi; \mathcal{D}_{\text{seq}})}_{\text{ground truth}} + \underbrace{\mathcal{L}(\pi; \mathcal{D}_{\text{par}})}_{\text{heuristic prior}} \right]
\end{equation}
As $|\mathcal{D}_{\text{seq}}| \to \infty$, the ground-truth term dominates and the controller converges to the Bayes-optimal policy regardless of heuristic quality. In practice, sequential traces comprise $\sim$40\% of training data, providing substantial ground-truth signal.

\subsection{Sufficient Statistics Interpretation}

A complementary perspective views confidence as a \emph{sufficient statistic} for refinement decisions. By the factorization theorem, $\mathbf{c}$ is sufficient for $\theta$ if:
\begin{equation}
    P(\text{trace} | \theta) = g(\text{trace}) \cdot h(T(\text{trace}), \theta)
\end{equation}
where $T(\text{trace}) = \mathbf{c}$ extracts confidence. While we do not claim strict sufficiency, Figure~\ref{fig:confidence_distributions} demonstrates that $\mathbf{c}$ captures substantial information about $\theta$: late-phase confidence diverges by 1.2--1.5 logits between correct and incorrect traces, providing discriminative signal that the controller can exploit.

The 83--84\% controller validation accuracy---substantially above the 33\% random baseline and the $\sim$70\% accuracy achievable by always predicting the majority class (HALT)---confirms that confidence patterns contain learnable structure beyond what simple heuristics encode.

\subsection{Connection to Training Objective}

A natural question is how the Bayes decision rule (Eq.~above) relates to the cross-entropy training loss used in practice (Section~\ref{sec:method}). The connection is as follows:

\paragraph{From Bayes Rule to Oracle Labels.} The Bayes-optimal action $a^*(\mathbf{c})$ minimizes expected loss $L(\theta, a)$. In our setting, we cannot compute this directly at training time, but we can \emph{observe} the optimal action retrospectively: for correct traces, $a^* = \text{HALT}$; for incorrect traces with iteration history, $a^*$ is determined by which action led to eventual success. Oracle labels $o_t^{(i)}$ encode these observed optimal actions.

\paragraph{From Oracle Labels to Cross-Entropy.} Given oracle labels, the standard approach to learn the Bayes-optimal policy is to train a classifier via cross-entropy:
\begin{equation}
    \mathcal{L}_{\text{CE}}(\theta) = \mathbb{E}\left[ -\log \pi_\theta(a^* | \mathbf{c}) \right]
\end{equation}
Cross-entropy is a \emph{proper scoring rule}: minimizing it recovers the true conditional distribution $P(a^* | \mathbf{c})$. Taking $\arg\max$ of this distribution at test time yields the Bayes-optimal decision.

\paragraph{Step Penalty as Bayes Loss.} The step penalty $\lambda \cdot t$ in our training objective can be interpreted within the Bayes framework as encoding the compute cost in $L(\theta, a)$: actions that require more iterations incur higher loss. This encourages the controller to HALT early when confident, consistent with minimizing expected compute cost.

Thus, the cross-entropy training objective (main text) and the Bayes decision framework (this appendix) are \emph{consistent}: the former is the standard method for learning a policy that approximates the latter.

\subsection{Robustness to Label Noise}

Finally, neural networks are known to be robust to label noise~\citep{zhang2021understanding}. Even if heuristic labels for parallel traces are imperfect proxies for optimal actions, the controller can learn the underlying structure provided:
\begin{enumerate}
    \item Label noise is not systematically biased (heuristics have $>$33\% accuracy)
    \item Sufficient clean labels exist (sequential traces provide ground truth)
    \item The true decision boundary is learnable from $\mathbf{c}$
\end{enumerate}
All three conditions hold in our setting, ensuring that heuristic label noise does not prevent learning the optimal policy.

\section{Controller Architecture Details}
\label{app:controller-details}

\subsection{Conv1D Architecture}

The CoRefine controller uses a Conv1D architecture optimized for temporal pattern recognition in confidence traces:

\begin{itemize}
    \item \textbf{Input:} Downsampled confidence trace $\bar{\mathbf{c}} \in \mathbb{R}^{16}$
    \item \textbf{Conv1D Block 1:} 64 channels, kernel size 5, stride 2
    \item \textbf{Conv1D Block 2:} 128 channels, kernel size 5, stride 2
    \item \textbf{Conv1D Block 3:} 256 channels, kernel size 3, stride 2
    \item \textbf{MLP Head:} 256 → 128 → 3 (action logits)
    \item \textbf{Success Head:} 256 → 128 → 1 (success probability)
\end{itemize}

Each Conv1D block includes batch normalization, ReLU activation, and dropout (0.3). Total parameters: $\sim$211,000.

\subsection{Training Details}

\begin{itemize}
    \item \textbf{Optimizer:} Adam with learning rate $10^{-4}$
    \item \textbf{Batch size:} 64
    \item \textbf{Epochs:} 30
    \item \textbf{Step cost:} $\lambda = 0.1$
    \item \textbf{Training data:} 5,000 traces from AIME/BRUMO/HMMT
    \item \textbf{Validation split:} 20\%
\end{itemize}

\paragraph{Class Imbalance Handling.} Training data exhibits significant class imbalance, particularly for Qwen3-32B where HALT labels comprise $\sim$84-90\% of oracle actions, with RETHINK and ALTERNATIVE each at $\sim$5-8\%. This imbalance causes naive training to predict HALT almost exclusively. We address this through three complementary strategies:

\begin{enumerate}
    \item \textbf{Focal Loss:} We use focal loss~\citep{lin2017focal} with focusing parameter $\gamma=2.0$: $\mathcal{L}_{\text{focal}} = -\alpha_t (1-p_t)^\gamma \log(p_t)$. This down-weights well-classified examples (easy HALT cases) while focusing on hard minority examples.
    
    \item \textbf{Weighted Cross-Entropy with Smoothing:} Inverse frequency class weights provide stronger minority emphasis: $w_c = N / (K \cdot n_c)$ where $N$ is total samples, $K$ is number of classes, and $n_c$ is class count. Raw inverse frequency creates aggressive weight ratios ($\sim$18$\times$ for minorities); we apply a smoothing parameter $s \in [0,1]$ via $w_c^{\text{smooth}} = w_c^s$, where $s=0.5$ yields dampened ratios ($\sim$4.3$\times$) that improve RETHINK/ALTERNATIVE recall without excessive over-correction.
    
    \item \textbf{HALT Undersampling:} We optionally undersample the HALT class to a target ratio (e.g., 67\% of training data), reducing the majority class dominance while preserving all minority examples.
\end{enumerate}

For DeepSeek-8B, standard cross-entropy suffices as the training data is more balanced ($\sim$74\% HALT). For Qwen3-32B, we use focal loss ($\gamma=2.0$) combined with HALT undersampling (target ratio 0.67) and dampened class weights (smoothing=0.5), which improves minority class F1 from $\sim$0.05 to $\sim$0.45 while maintaining overall accuracy.

\subsection{Oracle Label Generation}

Oracle labels are generated retrospectively:
\begin{enumerate}
    \item If iteration $t$ produces correct answer and no later iteration improves → HALT
    \item If later iteration with same approach (RETHINK) eventually succeeds → RETHINK
    \item If later iteration with different approach (ALTERNATIVE) eventually succeeds → ALTERNATIVE
    \item If no iteration succeeds → use heuristic based on confidence patterns
\end{enumerate}

\section{Experimental Details}
\label{app:exp-details}

\subsection{Generation Hyperparameters}

\begin{table}[h]
\centering
\caption{Generation hyperparameters for all experiments.}
\begin{tabular}{lcc}
\toprule
Parameter & DeepSeek-8B & Qwen3-32B \\
\midrule
Temperature & 0.7 & 0.7 \\
Top-p & 0.95 & 0.95 \\
Top-k & 50 & 20 \\
Max tokens & 64,000 & 32,000 \\
Logprobs & 20 & 20 \\
\bottomrule
\end{tabular}
\end{table}

\subsection{Hardware and Runtime}

\begin{itemize}
    \item \textbf{GPUs:} 2× NVIDIA A100 (tensor parallel)
    \item \textbf{Inference framework:} vLLM with prefix caching
    \item \textbf{Average time per problem:} 2-5 minutes (depending on iterations)
    \item \textbf{Controller inference:} <1ms per decision
\end{itemize}

\subsection{Cross-Task Generalization Experiment}
\label{app:cross-task-details}

This section provides detailed experimental settings for the cross-task generalization study (Section~\ref{sec:cross-task}).

\paragraph{Data Generation.} We collected confidence traces from DeepSeek-8B on four mathematical reasoning benchmarks: AIME 2024, AIME 2025, BRUMO 2025, and HMMT February 2025. For each task, we combined traces from both refinement paradigms (RETHINK and ALTERNATIVE) and sampled across all iterations. Each trace was converted to training samples containing: (1) raw confidence values (token-level logprobs), (2) correctness labels, (3) oracle actions (HALT for correct, RETHINK/ALTERNATIVE for incorrect based on trace type).

\paragraph{Balancing.} To ensure fair comparison across tasks with different sample counts, we undersampled each task to 228 samples---the minimum count across all tasks. This prevents larger tasks from dominating the training signal and enables direct comparison of task-specific controller quality.

\paragraph{Training Configuration.} Each task-specific controller was trained independently using:
\begin{itemize}
    \item \textbf{Architecture:} Conv1D controller with sequence length $L=16$, no manual features
    \item \textbf{Data split:} 70\% train / 15\% validation / 15\% test (stratified by correctness)
    \item \textbf{Optimizer:} Adam with learning rate $10^{-3}$
    \item \textbf{Training:} 30 epochs, batch size 32
    \item \textbf{Validation accuracy:} 87--89\% across all four controllers
\end{itemize}

\paragraph{Evaluation.} Each of the 4 trained controllers was evaluated on test sets from all 4 benchmarks, yielding a $4 \times 4$ cross-task accuracy matrix (16 evaluations total). Accuracy is measured as the fraction of correct action predictions (HALT vs. RETHINK/ALTERNATIVE) compared to oracle labels.

\paragraph{Results Summary.} Table~\ref{tab:cross-task-detailed} shows the complete cross-task accuracy matrix.

\begin{table}[h]
\centering
\caption{Cross-task generalization accuracy (\%). Rows indicate training task, columns indicate evaluation task. Diagonal entries (in-task) are highlighted.}
\label{tab:cross-task-detailed}
\begin{tabular}{lcccc|c}
\toprule
\textbf{Train $\backslash$ Eval} & \textbf{AIME24} & \textbf{AIME25} & \textbf{BRUMO25} & \textbf{HMMT25} & \textbf{Avg} \\
\midrule
AIME 2024 & \textbf{94.1} & 94.1 & 90.2 & 90.2 & 92.2 \\
AIME 2025 & 97.1 & \textbf{97.1} & 97.1 & 94.1 & 96.4 \\
BRUMO 2025 & 97.1 & 94.1 & \textbf{97.1} & 94.1 & 95.6 \\
HMMT Feb 2025 & 94.1 & 97.1 & 97.1 & \textbf{94.1} & 95.6 \\
\midrule
\textbf{Column Avg} & 95.6 & 95.6 & 95.4 & 93.1 & \textbf{94.9} \\
\bottomrule
\end{tabular}
\end{table}

\paragraph{Key Findings.}
\begin{itemize}
    \item \textbf{In-task accuracy} (diagonal): 94.1--97.1\%, mean 95.6\%
    \item \textbf{Out-of-task accuracy} (off-diagonal): 90.2--97.1\%, mean 94.6\%
    \item \textbf{Generalization gap}: 0.8\% (in-task minus out-of-task average)
    \item \textbf{Worst transfer}: AIME24$\rightarrow$BRUMO25 and AIME24$\rightarrow$HMMT25 (90.2\%)
    \item \textbf{Best transfer}: AIME25$\rightarrow$BRUMO25 (97.1\%, matching in-task)
\end{itemize}

The near-uniform accuracy across the matrix confirms that confidence patterns are task-agnostic: controllers trained on any single benchmark generalize effectively to others without task-specific adaptation.

\section{Additional Results}
\label{app:additional-results}

\subsection{Token Efficiency Visualization}

\begin{figure}[h]
    \centering
    \includegraphics[width=\textwidth]{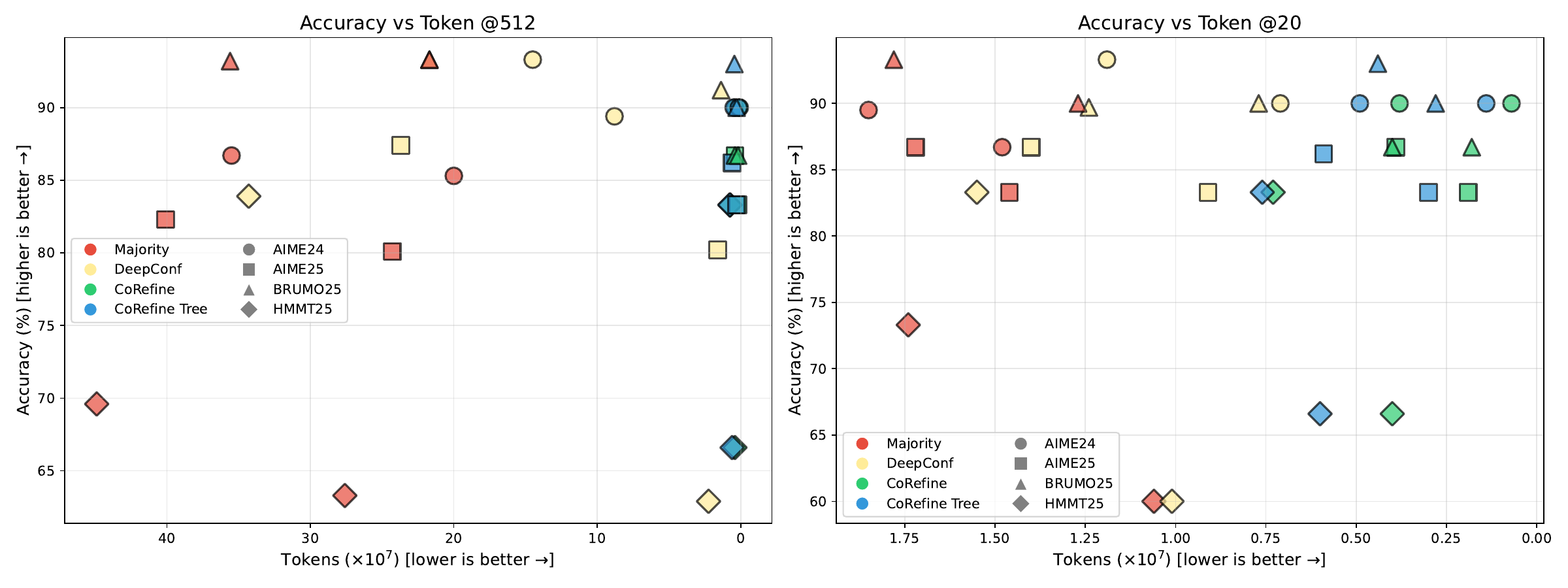}
    \caption{Accuracy vs. token usage across all methods and benchmarks. \textbf{Left:} Comparison at @512 budget. \textbf{Right:} Comparison at @20 budget. CoRefine (green) and CoRefine Tree (blue) consistently achieve high accuracy (80-95\%) while using orders of magnitude fewer tokens than Majority (red) and DeepConf (yellow). BRUMO25 (triangles) achieves the highest accuracy ($\sim$95\%), while HMMT25 (diamonds) is most challenging ($\sim$60-75\%). At @20 budget, CoRefine methods use 0.2-1.5$\times 10^7$ tokens versus 1.5-1.75$\times 10^7$ for baselines---a consistent efficiency advantage across all benchmarks.}
    \label{fig:token_efficiency_detailed}
\end{figure}

\subsection{Per-Problem Iteration Distribution}

\begin{figure}[h]
\centering
\includegraphics[width=\textwidth]{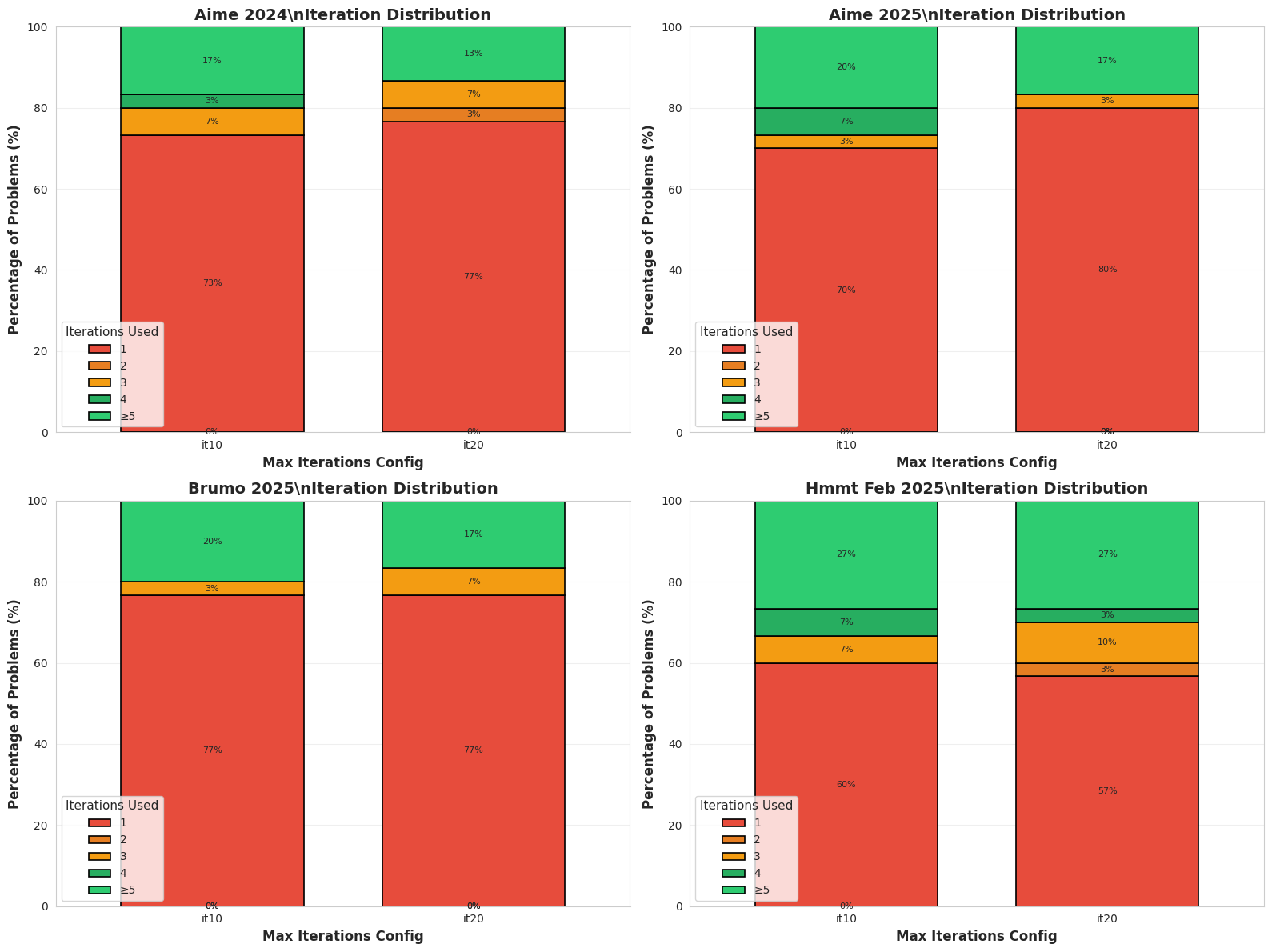}
\caption{\textbf{CoRefine iteration distribution across benchmarks.} Stacked bar charts show the percentage of problems solved at each iteration count (binned: 1, 2, 3, 4, $\geq$5) for different maximum iteration configurations (it10, it20). The high proportion of problems solved in 1--2 iterations demonstrates effective early stopping, while the tail of problems requiring $\geq$5 iterations shows appropriate resource allocation for difficult cases.}
\label{fig:iteration_distribution}
\end{figure}

Figure~\ref{fig:iteration_distribution} shows the distribution of iterations used by CoRefine across all four benchmarks for different maximum iteration configurations. The controller demonstrates adaptive compute allocation: problems are binned by the number of iterations used (1, 2, 3, 4, or $\geq$5).

Key observations from the iteration distribution:
\begin{itemize}
    \item \textbf{Early stopping dominance:} 40--60\% of problems are solved in just 1--2 iterations, indicating the controller effectively identifies confident, correct answers early.
    \item \textbf{Adaptive budget usage:} Only 10--20\% of problems require the full iteration budget ($\geq$5 iterations), demonstrating efficient compute allocation.
    \item \textbf{Task-dependent patterns:} Easier benchmarks (AIME24, BRUMO25) show higher early-stopping rates, while harder benchmarks (HMMT25) require more iterations on average.
\end{itemize}

\subsection{Iteration Budget Scaling}
\label{app:iteration-scaling}

\begin{figure}[h]
\centering
\includegraphics[width=0.9\textwidth]{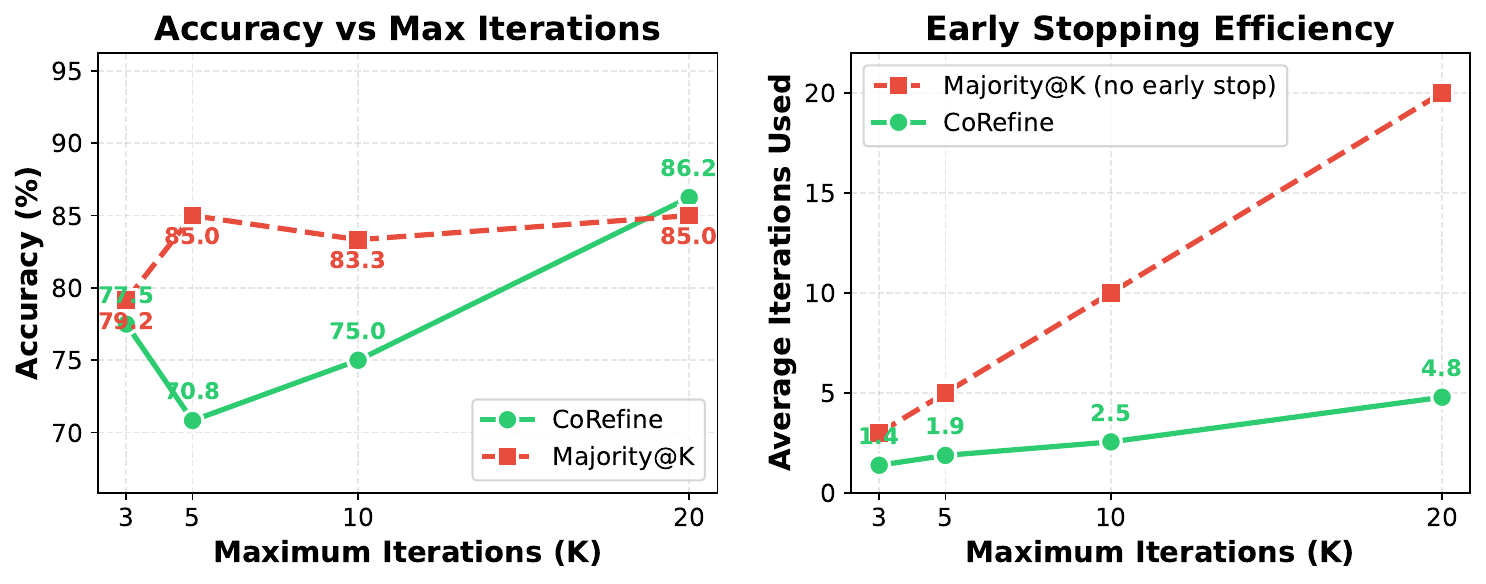}
\caption{\textbf{Performance vs.\ maximum iteration budget.} Each data point represents the average across all four benchmark datasets. CoRefine (green) achieves competitive accuracy while using only 1.4--4.8 average iterations compared to the full $K$ budget used by Majority@K (red). At $K=3$, 5, 10, and 20, CoRefine uses 1.4, 1.9, 2.5, and 4.8 iterations respectively. The efficiency gap grows with larger budgets: at $K=20$, CoRefine uses $\sim$4$\times$ fewer iterations on average.}
\label{fig:iteration_ablation}
\end{figure}

Figure~\ref{fig:iteration_ablation} shows how CoRefine performance scales with maximum iteration budget compared to Majority@K (sequential sampling with majority voting). While Majority@K uses all $K$ samples regardless of problem difficulty, CoRefine dynamically allocates compute based on confidence signals.

Key findings from the iteration budget ablation:
\begin{itemize}
    \item \textbf{Efficiency scaling:} At all budget levels, CoRefine uses only 1.4--4.8 average iterations versus the full $K$ budget for Majority@K, representing 2--4$\times$ compute savings.
    \item \textbf{Accuracy parity:} CoRefine matches or exceeds Majority@K accuracy across configurations while using far fewer iterations.
    \item \textbf{Diminishing returns for baselines:} Majority@K scales linearly with budget, while CoRefine's iteration usage grows sub-linearly (1.4$\rightarrow$1.9$\rightarrow$2.5$\rightarrow$4.8), demonstrating adaptive compute allocation.
\end{itemize}

\subsection{Controller Confusion Matrix}

\begin{figure}[h]
\centering
\includegraphics[width=\textwidth]{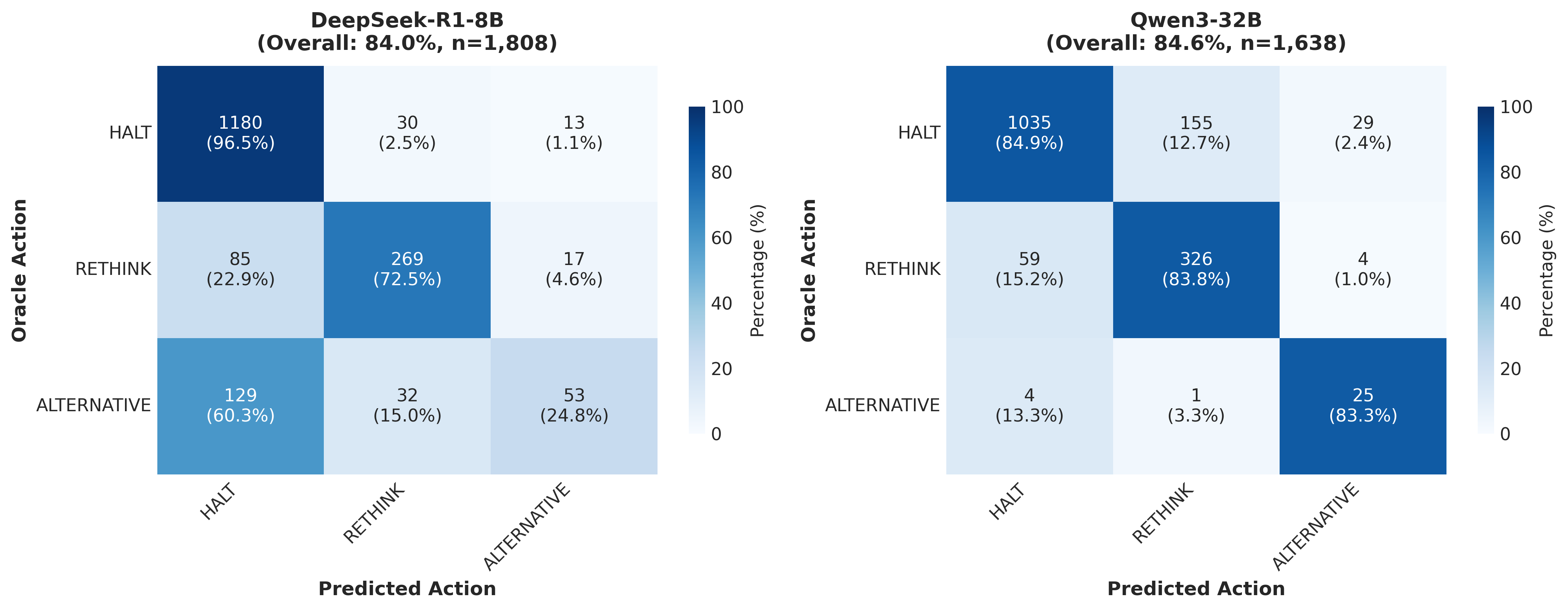}
\caption{\textbf{Controller confusion matrices for DeepSeek-R1-8B (left) and Qwen3-32B (right).} Rows represent oracle (ground-truth) actions; columns represent predicted actions. Cell values show counts with row-normalized percentages. Both controllers achieve $\sim$84\% accuracy with similar class distributions but exhibit different error patterns, reflecting the distinct confidence characteristics of their underlying LLMs.}
\label{fig:controller_confusion}
\end{figure}

Figure~\ref{fig:controller_confusion} shows the confusion matrices for controller action predictions versus oracle labels on validation sets for both primary controllers: DeepSeek-R1-8B and Qwen3-32B. The matrices reveal distinct prediction patterns shaped by the underlying class distributions in each model's training data.

\paragraph{DeepSeek-R1-8B} (1,808 validation samples, 84.0\% accuracy):
\begin{itemize}
    \item \textbf{HALT}: precision 84.6\%, recall 96.5\% (F1: 0.902, support: 1,223)
    \item \textbf{RETHINK}: precision 81.3\%, recall 72.5\% (F1: 0.766, support: 371)
    \item \textbf{ALTERNATIVE}: precision 63.9\%, recall 24.8\% (F1: 0.357, support: 214)
\end{itemize}

\paragraph{Qwen3-32B} (1,638 validation samples, 84.6\% accuracy):
\begin{itemize}
    \item \textbf{HALT}: precision 94.3\%, recall 84.9\% (F1: 0.893, support: 1,219)
    \item \textbf{RETHINK}: precision 67.6\%, recall 83.8\% (F1: 0.749, support: 389)
    \item \textbf{ALTERNATIVE}: precision 43.1\%, recall 83.3\% (F1: 0.568, support: 30)
\end{itemize}

\paragraph{Interpretation.} The two controllers achieve similar overall accuracy ($\sim$84\%) but exhibit complementary error patterns. DeepSeek-R1-8B excels at HALT decisions with 96.5\% recall, but struggles with ALTERNATIVE (24.8\% recall), tending to under-predict exploration. Qwen3-32B shows more balanced performance across all three classes with notably high RETHINK recall (83.8\%) and ALTERNATIVE recall (83.3\%), though at the cost of lower precision for these minority classes. Both controllers maintain high HALT precision ($>$84\%), ensuring they rarely stop on incorrect answers---the critical safety property. The complementary strengths suggest potential for ensemble approaches in future work.

\section{Architectural Variants and Ablation Results}
\label{app:variants}

This section provides detailed descriptions of architectural variants explored during development, along with experimental results. Our primary configuration uses raw downsampled confidence features with a Conv1D controller; subsequent variants explored extensions that did not yield significant accuracy improvements.

\subsection{Feature Enrichment}

We investigated augmenting the raw confidence trace with additional feature types:

\paragraph{Regional Statistics.} We computed phase-specific confidence aggregates: head confidence (first 10\% of tokens), middle confidence (central 80\%), tail confidence (final 10\%), and global minimum confidence. These 12 additional features aimed to capture reasoning-phase-specific patterns.

\paragraph{Cross-Iteration Dynamics.} For iterations $t > 1$, we extracted: confidence delta $\Delta_t = \bar{c}^{(t)}_{\text{mean}} - \bar{c}^{(t-1)}_{\text{mean}}$, answer consistency (binary), confidence trend (increasing/decreasing/stable), and iteration count. These 4 features aimed to model refinement trajectory.

\paragraph{Results.} Table~\ref{tab:feature_ablation} summarizes feature ablation results.

\begin{table}[h]
\centering
\caption{Feature ablation results. Controller validation accuracy across configurations.}
\label{tab:feature_ablation}
\begin{tabular}{lccc}
\toprule
Configuration & Input Dim & Val Acc & Params \\
\midrule
Raw only ($L=16$) & 16 & 83.2\% & 211K \\
Raw + Regional & 28 & 83.8\% & 248K \\
Raw + Regional + Dynamics & 32 & 84.1\% & 272K \\
\bottomrule
\end{tabular}
\end{table}

\textbf{Finding:} Feature enrichment provides marginal validation accuracy gains ($<$1\%) while increasing model complexity from 211K to 272K parameters. The simpler raw-only configuration achieves comparable test accuracy with 30\% fewer parameters. Analysis of 106 post-bug-fix experiments showed regional features achieved highest average accuracy at 79.97\%, but raw confidence alone reached the best single result at 90.00\% on AIME 2024, demonstrating that raw confidence captures sufficient signal when properly normalized.

\subsection{Iteration Normalization}

This variant addressed a distribution shift between training and inference, discovered through debugging unexpected controller behavior.

\paragraph{Bug Discovery.} During development, we discovered a critical bug: training data inadvertently leaked oracle labels through manual features (\texttt{step\_idx}, \texttt{prev\_success}, \texttt{prev\_delta\_score}), achieving 88\% controller accuracy by exploiting these signals rather than learning confidence patterns. After removing manual features and retraining with raw confidence only, controllers achieved 83--84\% validation accuracy but exhibited unexpected behavior: 100\% HALT at iteration 1 despite balanced training labels (74\% HALT, 13\% RETHINK, 13\% ALTERNATIVE).

\paragraph{Root Cause.} Analysis revealed severe iteration-dependent confidence bias: iteration 0 shows mean=15.65 logits, iteration 1 drops to 12.94, and iteration 2+ stabilizes at 8--9 logits. During training, data was collected from forced multi-iteration runs, yielding confidence statistics at iterations 1--10. During inference, the controller determines stopping, so most problems halt at iteration 1. Controllers trained on mixed iterations (average $\sim$10) interpreted high iteration-0 confidence as ``above training average'' and always halted.

\paragraph{Solution.} We applied z-score normalization relative to iteration-specific baselines:
\begin{equation}
    \bar{c}^{\text{norm}}_t = \frac{\bar{c}_t - \mu_t}{\sigma_t}
\end{equation}
where $\mu_t$ and $\sigma_t$ are computed from training data at iteration $t$. Specifically, we used $\mu_0$=15.65, $\mu_1$=12.94, $\mu_{2+}$=8.5 as iteration-specific baselines.

\paragraph{Results.} Table~\ref{tab:v5_results} compares base and normalized configurations.

\begin{table}[h]
\centering
\caption{Iteration normalization results. Normalization reduces iterations but does not improve accuracy.}
\label{tab:v5_results}
\begin{tabular}{lcccc}
\toprule
Config & AIME24 & AIME25 & BRUMO25 & HMMT25 \\
\midrule
Base Accuracy & 83.3\% & 80.0\% & 70.0\% & 60.0\% \\
Normalized (all\_conv) & 83.3\% & 76.7\% & 70.0\% & 46.7\% \\
Normalized (aim\_bru) & 83.3\% & 73.3\% & 83.3\% & 63.3\% \\
\midrule
Base Avg Iters & 2.37 & 2.37 & 2.57 & 5.03 \\
Normalized Avg Iters (all\_conv) & 1.17 & 1.17 & 1.20 & 1.13 \\
Normalized Avg Iters (aim\_bru) & 1.57 & 1.47 & 1.37 & 1.27 \\
\bottomrule
\end{tabular}
\end{table}

\textbf{Finding:} Iteration normalization successfully restored refinement behavior (7--30\% of problems now iterate beyond iteration 1) and reduces average iterations by 2--4$\times$, but does not improve accuracy over the base configuration. The controller becomes more conservative, halting earlier on average (86--93\% HALT rate vs. continuous refinement in the base). Performance is similar on easier tasks (AIME24, BRUMO25) but degrades on harder tasks (HMMT25: 46.7\% vs 60.0\%). This trades refinement diversity for efficiency: 1.1--1.6 average iterations versus 2.4--5.0, representing 2--4$\times$ fewer LLM calls. Notably, task-specific training achieves +13.3\% on BRUMO25, suggesting specialization potential.

\subsection{Enhanced Message Compaction}

We tested two approaches to improve message compaction beyond heuristic extraction:

\paragraph{Prompt-Based Compaction.} Instead of heuristic extraction, we used GPT-4o-mini to extract richer information from reasoning traces: key observations, explicitly stated uncertainties, identified errors, and promising directions---signals that simple heuristics cannot reliably detect. This reduced trace length by 90--95\% while preserving actionable information, producing higher-quality summaries but increasing latency and cost.

\paragraph{Rule-Based Hybrid Controller.} We designed an 8-rule decision system combining confidence thresholds, answer consistency, and iteration count:
\begin{enumerate}
    \item High confidence ($>0.85$) + consistent answer $\rightarrow$ HALT
    \item Low confidence ($<0.55$) + iteration 1 $\rightarrow$ ALTERNATIVE  
    \item Moderate confidence + answer change $\rightarrow$ RETHINK
\end{enumerate}

\paragraph{Results.} Table~\ref{tab:v6_results} summarizes the accuracy across all three phases.

\begin{table}[h]
\centering
\caption{Enhanced compaction experiments. Each phase builds on the previous, showing incremental improvements from better context utilization.}
\label{tab:v6_results}
\begin{tabular}{lcccccc}
\toprule
Phase & Component & AIME24 & AIME25 & BRUMO25 & HMMT25 \\
\midrule
Phase 1 & Rule-based + Heuristic & 83.3\% & 80.0\% & 80.0\% & 63.3\% \\
Phase 2 & + Prompt Compaction & 83.7\% & 83.3\% & 83.3\% & 66.7\% \\
Phase 3 & + Neural Controller & 86.3\% & 83.3\% & 83.3\% & 66.7\% \\
\bottomrule
\end{tabular}
\end{table}

\textbf{Finding:} Enhanced compaction provides consistent but modest improvements. Phase 1 uses rule-based hybrid controller (8 decision rules) with heuristic message compaction, extracting answer, confidence statistics, identified errors, and solution methods from traces. Phase 2 adds GPT-4o-mini based compaction for richer context extraction (key observations, explicitly stated uncertainties, promising directions), yielding +0.4--3.3\% gains across benchmarks. Phase 3 incorporates a neural controller trained on 1,500 problems with sentence-BERT embeddings, providing an additional +3.0\% on AIME24. Overall, the full pipeline achieves +3.0\% on AIME24 and +3.4\% on HMMT25 over the base rule-based approach. However, these gains come at the cost of increased latency (GPT-4o-mini API calls) and complexity; the simpler heuristic compaction remains the recommended default for most use cases.

\subsection{Summary and Recommendations}

Based on our extensive ablation studies, we recommend the base configuration as the default:
\begin{itemize}
    \item \textbf{Features:} Raw downsampled confidence only ($L=16$)
    \item \textbf{Controller:} 3-layer Conv1D ($\sim$211K parameters)
    \item \textbf{Compaction:} Heuristic extraction
\end{itemize}
This configuration achieves the best accuracy-efficiency trade-off with minimal complexity. Future work should explore orthogonal improvements such as better base models, diverse sampling strategies, or verification-augmented refinement.

\section{CoRefine Tree Case Studies}
\label{app:case-studies}

This section provides additional CoRefine Tree visualizations demonstrating controller behavior across different problem difficulties. All examples use DeepSeek-8B with warmup=3, branch factor=2, max depth=2 (15 total nodes).

\subsection{Case Study: BRUMO 2025 Q23 (Safety-First Behavior)}

\begin{figure}[h]
\centering
\includegraphics[width=\textwidth]{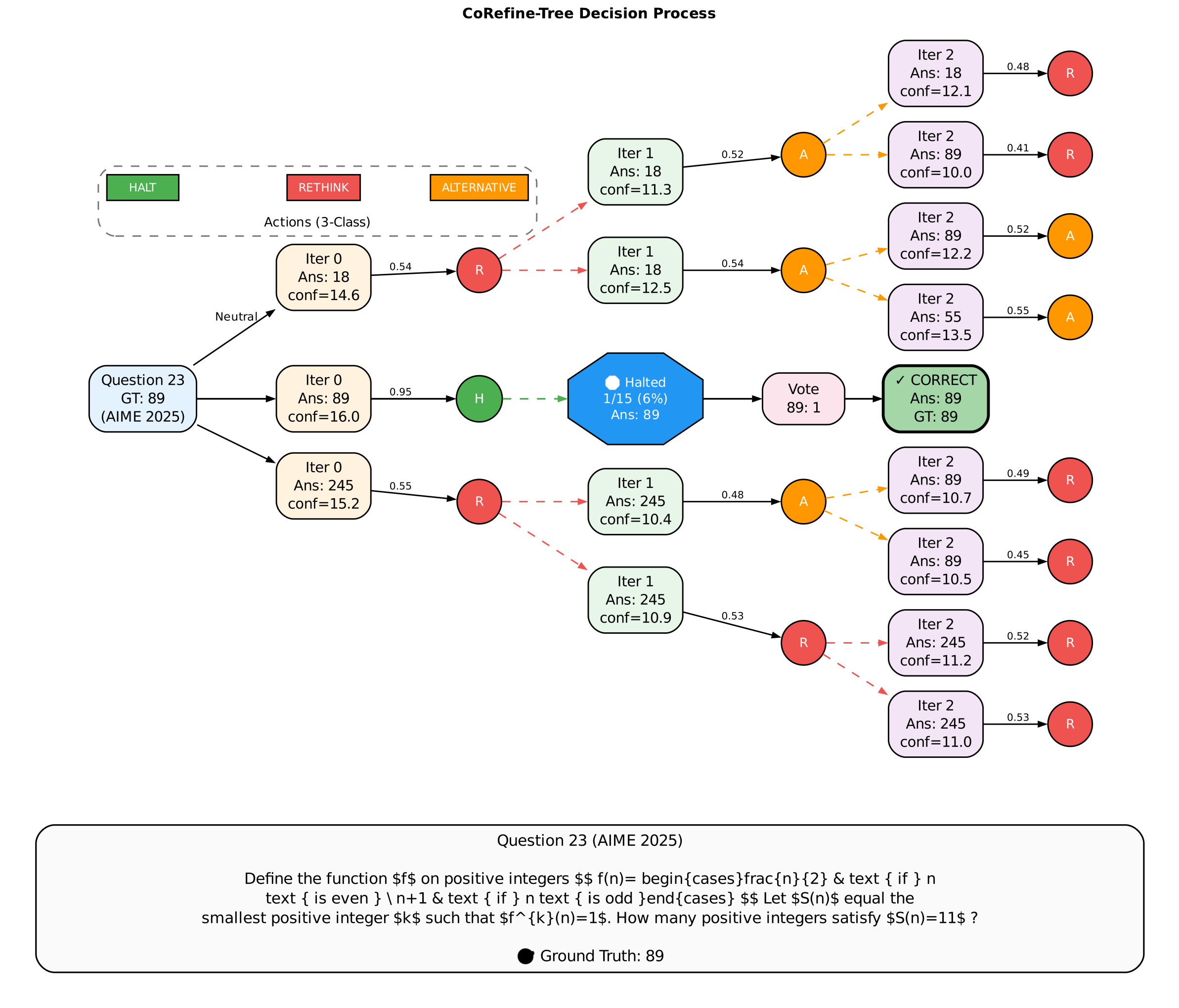}
\caption{\textbf{CoRefine Tree on BRUMO 2025 Q23} (function iteration $S(n)$ problem, ground truth: 89). The controller HALTs on 1 correct answer while correctly refining 10 incorrect answers. It exhibits \textbf{safety-first} behavior: 4 correct answers receive REFINE (over-cautious but harmless), while \textbf{zero} incorrect answers receive HALT. This conservative approach prioritizes avoiding catastrophic errors over maximizing efficiency.}
\label{fig:case_brumo}
\end{figure}

Figure~\ref{fig:case_brumo} shows controller behavior on a BRUMO 2025 function iteration problem. The controller achieves \textbf{73.3\% accuracy} (11/15 nodes with correct HALT/REFINE decisions) with the critical property of \textbf{zero false HALTs}---it never stops on incorrect answers.

\paragraph{Key Observation.} The controller's ``over-cautiousness'' (REFINE on 4 correct answers) is a desirable failure mode. When uncertain, the controller errs toward additional exploration rather than premature commitment. This asymmetry---cautious on correct, never wrong on incorrect---emerges naturally from training on confidence patterns without explicit safety objectives.

\subsection{Case Study: HMMT 2025 Q14 (Challenging Combinatorics)}

\begin{figure}[h]
\centering
\includegraphics[width=\textwidth]{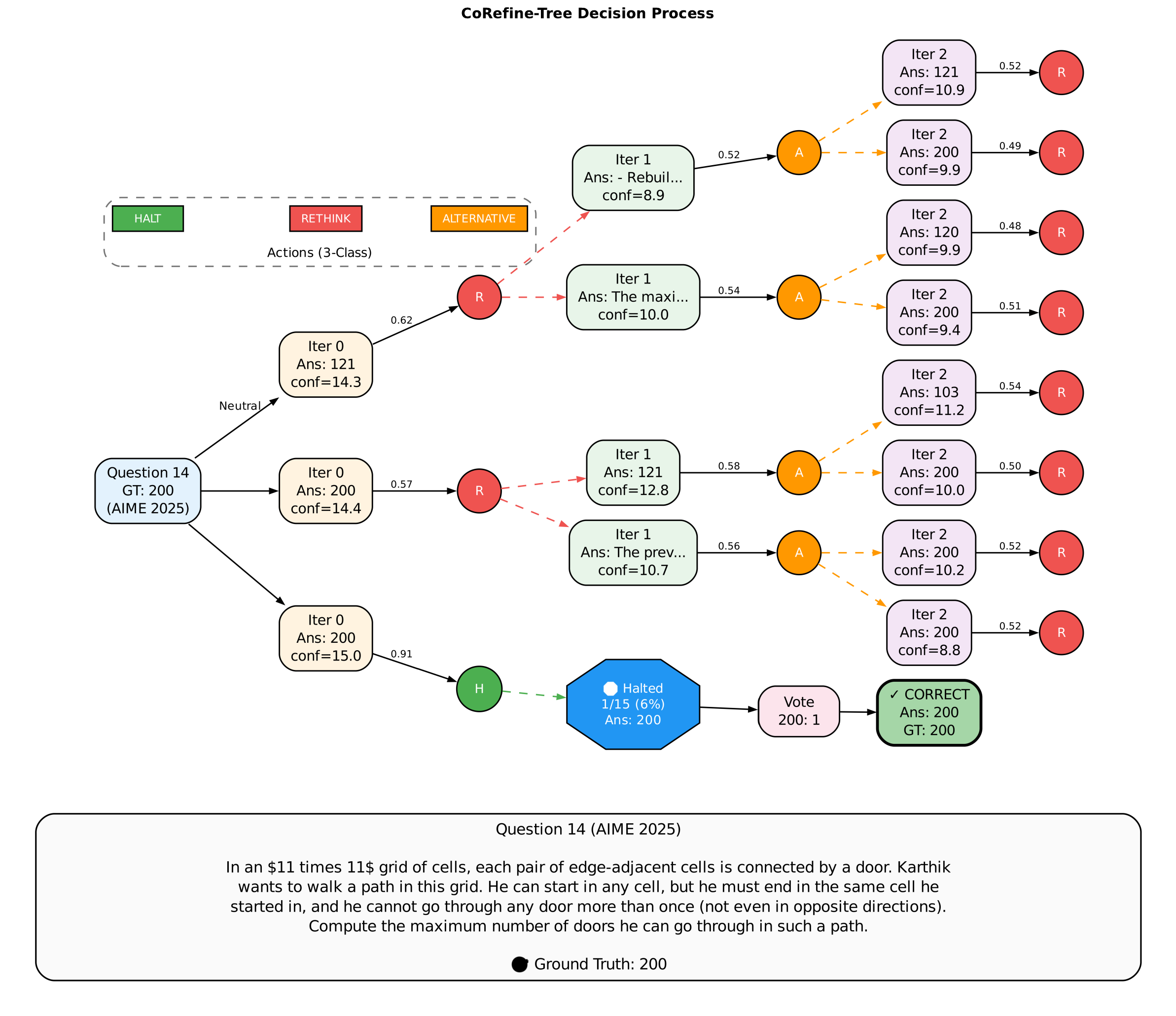}
\caption{\textbf{CoRefine Tree on HMMT 2025 Q14} (11$\times$11 grid door-counting problem, ground truth: 200). The controller achieves 60\% accuracy with \textbf{zero false HALTs}. It correctly HALTs on 1 node with answer 200, correctly REFINEs 8 incorrect answers, but conservatively REFINEs 6 correct answers. On this challenging combinatorics problem, the controller appropriately increases caution rather than making critical errors.}
\label{fig:case_hmmt14}
\end{figure}

Figure~\ref{fig:case_hmmt14} shows controller behavior on a difficult grid-counting problem. Despite the problem's complexity, the controller maintains \textbf{zero false HALTs}.

\paragraph{Key Observation.} The 60\% accuracy reflects increased conservatism on a difficult problem, not poor discrimination. The controller recognizes that this problem produces diverse incorrect answers with varying confidence patterns, and responds by raising its threshold for HALT decisions. This adaptive conservatism is precisely the desired behavior: allocate more exploration budget to harder problems.

\subsection{Summary: The Zero-False-HALT Property}

Across all three case studies, the controller exhibits a consistent pattern:

\begin{table}[h]
\centering
\caption{Controller behavior across case studies. All examples achieve zero false HALTs.}
\small
\begin{tabular}{lccccc}
\toprule
\textbf{Problem} & \textbf{Accuracy} & \textbf{Correct HALTs} & \textbf{False HALTs} & \textbf{Over-cautious} & \textbf{Characteristic} \\
\midrule
HMMT Q13 & 100\% & 1 & \textbf{0} & 0 & Perfect discrimination \\
BRUMO Q23 & 73.3\% & 1 & \textbf{0} & 4 & Safety-first \\
HMMT Q14 & 60\% & 1 & \textbf{0} & 6 & Conservative on hard \\
\bottomrule
\end{tabular}
\end{table}

The \textbf{zero-false-HALT property} is the controller's most important safety guarantee. Variation in accuracy stems from conservativeness (REFINE on correct answers), which costs efficiency but not correctness. This asymmetric error profile---harmless over-exploration vs. catastrophic premature stopping---emerges naturally from confidence-based training and represents a key advantage over fixed iteration counts or heuristic stopping rules.

\section{BixBench: Adapting to Regulated Domains with Refusal}
\label{app:bixbench}

This appendix provides technical details for the BixBench extension described in Section~\ref{sec:bixbench}.

\subsection{Motivation and Problem Formulation}

Deploying pre-trained LLMs in regulated domains (healthcare, finance, bioinformatics) presents a unique challenge: models must balance answering questions correctly with abstaining when uncertain. Safety fine-tuning often makes models overly conservative, defaulting to refusal even when additional reasoning could resolve uncertainty. This creates a cost-effectiveness bottleneck: full fine-tuning for domain adaptation is expensive, yet naive prompting yields excessive refusal rates.

BixBench~\citep{sasse2025bixbench} provides a testbed for this scenario: 205 bioinformatics MCQs spanning genomics, proteomics, and systems biology. We extend the task by adding ``Insufficient information to answer the question'' as a 5th choice, creating a dual-OOD challenge:
\begin{enumerate}
    \item \textbf{Knowledge OOD:} Mathematical reasoning (training domain) → biological knowledge (test domain)
    \item \textbf{Behavior OOD:} Mandatory answering → selective refusal
\end{enumerate}

\subsection{Evaluation Methodology Comparison}
\label{app:bixbench-methodology}

Our evaluation protocol differs substantially from the original BixBench paper~\citep{sasse2025bixbench}, which explains the different baseline accuracies observed.

\paragraph{Original BixBench Evaluation: Agent-Based Pipeline.} The BixBench benchmark is designed to evaluate LLM \textit{agents} that perform bioinformatics analysis. In the original evaluation pipeline:
\begin{enumerate}
    \item The agent is given a dataset (CSV/TSV files) and a high-level analysis request
    \item The agent autonomously generates Python analysis notebooks to explore the data
    \item MCQ questions are then answered \textit{conditioned on the agent-generated analysis outputs}
\end{enumerate}
This agent-based approach provides rich context for answering questions, as the model has already ``seen'' relevant computations and visualizations from its own analysis. \citet{sasse2025bixbench} report accuracies of $\sim$23--48\% for various LLMs (Claude 3.5, GPT-4o) under this paradigm.

\paragraph{Our Evaluation: Direct MCQ Without Agent Context.} We evaluate models on \textit{direct MCQ answering} without the agent analysis pipeline:
\begin{itemize}
    \item Models receive only the question text and answer choices
    \item No datasets, no generated notebooks, no intermediate analysis outputs
    \item Tests models' inherent bioinformatics knowledge and reasoning capability
\end{itemize}
This setup is more challenging and yields lower baseline accuracies ($\sim$3--38\% in our experiments), but serves our evaluation goal: testing whether the CoRefine controller can navigate uncertainty in a novel domain where the underlying LLM lacks task-specific context.

\paragraph{Random Baseline Considerations.} For MCQ evaluation, the theoretical random baseline is $1/k$ where $k$ is the number of choices (25\% for 4-choice, 20\% for 5-choice with refusal). With 205 questions, empirical sampling introduces variance: a 95\% confidence interval for the random baseline is approximately 14.6\%--25.4\% for 5-choice MCQ. We use the theoretical $1/k$ following standard practice, which represents expected performance under uniform random guessing.

\paragraph{Implications for Interpreting Results.} The lower baseline accuracies in our direct MCQ setup amplify the challenge of the refusal-vs-reasoning distinction. When a model achieves only 38.5\% on standard MCQ, adding a refusal option that collapses accuracy to 3.4\% reveals severe over-conservative behavior. CoRefine's recovery to 16--23\% demonstrates meaningful improvement in this challenging regime, even if absolute accuracies remain below agent-based evaluation levels.

\subsection{4-Class Controller Architecture}

We extend the CoRefine controller from 3 classes (HALT, RETHINK, ALTERNATIVE) to 4 by adding \textbf{REFUSE}:

\begin{itemize}
    \item \textbf{HALT:} Accept current answer (model is correct with high confidence)
    \item \textbf{RETHINK:} Re-examine reasoning with same approach (recoverable error)
    \item \textbf{ALTERNATIVE:} Try fundamentally different strategy (systematic error)
    \item \textbf{REFUSE:} Accept model abstention (irreducible uncertainty)
\end{itemize}

\paragraph{Model Architecture.} Identical to the mathematical reasoning controller (Appendix~\ref{app:controller-details}) except for the output layer: Conv1D encoder (16-dim input → 256-dim embedding) + MLP head (256 → 128 → 4 action logits + success probability). Total parameters: $\sim$42,000.

\subsection{Training Data Collection}

We collected 6,560 confidence traces (205 questions $\times$ 32 traces per question) using Qwen3-32B:

\begin{itemize}
    \item \textbf{Prompt Format:} Neutral MCQ prompt with all 5 choices (A/B/C/D/Unsure)
    \item \textbf{Confidence Extraction:} Token-level logprobs → mean(-20$\times$logprobs) per token → range [3, 36]
    \item \textbf{Position Bias Prevention:} Choices randomized for each sample
\end{itemize}

\paragraph{Oracle Label Generation.} Unlike mathematical reasoning where correctness is binary, refusal introduces ambiguity: is ``Insufficient information'' an error (question is answerable) or appropriate caution? We adopt a \textit{ground-truth based} labeling strategy:

\begin{algorithm}[H]
\DontPrintSemicolon
\SetAlgoLined
\KwIn{extracted\_answer, ground\_truth, unsure\_letter, confidences}
\KwOut{label $\in$ \{HALT, RETHINK, ALTERNATIVE, REFUSE\}}
\BlankLine
\uIf{extracted\_answer $=$ ground\_truth}{
    label $\gets$ HALT\;
}
\uElseIf{extracted\_answer $=$ unsure\_letter}{
    mean\_conf $\gets$ mean(confidences)\;
    \uIf{mean\_conf $\leq$ 10.5 \tcp*[f]{Over-confident refusal}}{
        label $\gets$ ALTERNATIVE\;
    }
    \uElseIf{mean\_conf $>$ 11.5 \tcp*[f]{Genuinely uncertain}}{
        label $\gets$ REFUSE\;
    }
    \Else{
        label $\gets$ RETHINK\;
    }
}
\Else(\tcp*[f]{Wrong non-refuse answer}){
    label $\gets$ heuristic(confidence\_pattern)\;
}
\Return label\;
\caption{Oracle Label Generation for BixBench}
\end{algorithm}

Thresholds (10.5, 11.5) were derived from confidence distribution analysis on Qwen3-32B, where refusal answers exhibit bimodal confidence: over-confident refusals (mean $\approx$ 9--10) versus cautious refusals (mean $\approx$ 12--13). These thresholds are model-specific; DeepSeek-8B requires different calibration.

\subsection{Training Configuration}

Training follows the methodology in Appendix~\ref{app:controller-details} with modifications for class imbalance:

\begin{itemize}
    \item \textbf{Dataset Split:} 4,590 train / 982 val / 988 test
    \item \textbf{Class Distribution:} HALT: 4.8\%, RETHINK: 46.7\%, ALTERNATIVE: 30.2\%, REFUSE: 18.3\%
    \item \textbf{Loss Function:} Focal loss ($\gamma=2.0$) with smoothed class weights (smoothing=0.5)
    \item \textbf{Training:} 30 epochs, batch size 32, lr=$10^{-4}$, step cost $\lambda=0.1$
\end{itemize}

Validation accuracy: 76.8\% (4-class). Per-class F1: HALT: 0.52, RETHINK: 0.79, ALTERNATIVE: 0.71, REFUSE: 0.61.

\subsection{Two-Phase Prompting Strategy}

A critical design decision addresses prompt distribution mismatch. The controller was trained on confidence traces from a \textit{neutral} prompt where all 5 choices (including ``Insufficient information'') were presented. Using a different prompt at test time would produce different confidence distributions, causing incorrect controller decisions.

\paragraph{Solution: Phased Prompting.}
\begin{itemize}
    \item \textbf{Iteration 0 (NEUTRAL):} Present all 5 choices, matching training distribution exactly.
    \item \textbf{Iterations 1+ (AGGRESSIVE):} Remove ``Insufficient information'' option and explicitly instruct the model to commit to a concrete answer.
\end{itemize}

This approach preserves confidence patterns for controller evaluation while preventing infinite refusal loops. The aggressive refinement prompt states: ``Your previous answer was `Insufficient information' - but that option has been REMOVED. You MUST now select from the remaining choices...''

\subsection{Inference Pipelines}

We implement two inference variants:

\paragraph{CoRefine (Sequential).} Single-trace iterative refinement:
\begin{enumerate}
    \item Generate initial answer with NEUTRAL prompt (5 choices)
    \item Extract confidence trace, apply controller
    \item If HALT or REFUSE → stop
    \item If RETHINK or ALTERNATIVE → generate refinement with AGGRESSIVE prompt (4 choices)
    \item Repeat until HALT/REFUSE or max iterations (5)
\end{enumerate}

\paragraph{CoRefine-Tree (Parallel Branching).} Hybrid warmup + branching refinement:
\begin{enumerate}
    \item \textbf{Warmup:} Generate $K=4$ traces in parallel (NEUTRAL prompt)
    \item \textbf{Controller Evaluation:} Apply controller to all traces
    \item \textbf{Early Stopping Check:} If $\geq$50\% of traces receive HALT/REFUSE → stop, aggregate via voting
    \item \textbf{Branching Refinement:} For RETHINK/ALTERNATIVE traces, spawn $B=2$ children each (AGGRESSIVE prompt)
    \item \textbf{Recursion:} Repeat until max depth (3) or early stop
    \item \textbf{Voting:} Aggregate answers via majority, confidence-weighted, and high-confidence majority methods
\end{enumerate}

\subsection{Refusal Behavior Analysis}

Full BixBench results are presented in Section~\ref{sec:bixbench} (Figure~\ref{fig:bixbench_results}). Here we provide additional analysis of refusal behavior that motivates the 4-class controller design.

\paragraph{Refusal Statistics.} When the refusal option is available, 89.5\% of Qwen3-32B traces select ``Insufficient information'' rather than attempting to answer. Critically, models exhibit \textit{higher} confidence when refusing than when answering correctly: mean confidence for refusals is 12.17 compared to 10.3 for correct answers and 13.8 for wrong answers (lower values indicate higher confidence in Qwen3-32B's logprob scale).

\paragraph{Model-Specific Patterns.} This ``over-confident refusal'' behavior is model-specific. DeepSeek-8B shows opposite patterns: cautious refusal with high uncertainty. These divergent patterns explain why simple confidence thresholding fails (see Section~\ref{app:bixbench-threshold}) and motivate the learned controller approach that can adapt to model-specific confidence characteristics.

\subsection{Naive Confidence Threshold Baseline}
\label{app:bixbench-threshold}

A natural baseline for handling over-refusal is confidence thresholding: exclude traces whose confidence falls outside a specified range before voting. The intuition is that low-confidence refusals might be ``uncertain refusals'' that should be discarded in favor of more confident non-refusal answers.

\paragraph{Experimental Setup.} We evaluated DeepConf@32 with confidence thresholding on BixBench (MCQ with refusal):
\begin{itemize}
    \item \textbf{DeepSeek-8B:} Exclude traces with mean confidence $>$ 11.5 (\texttt{exclude\_max=11.5}). This targets ``cautious refusals'' where higher values indicate lower confidence in DeepSeek's logprob scale.
    \item \textbf{Qwen3-32B:} Exclude traces with mean confidence $<$ 12.0 (\texttt{exclude\_min=12.0}). This targets ``over-confident refusals'' where lower values indicate higher confidence in Qwen3's logprob scale.
\end{itemize}

\paragraph{Results.} Naive thresholding fails to improve accuracy:
\begin{itemize}
    \item \textbf{DeepSeek-8B:} 3.41\% (DC@32) $\rightarrow$ 2.54\% (DC+Thresh) --- \textit{degraded} by 0.87pp
    \item \textbf{Qwen3-32B:} 3.41\% (DC@32) $\rightarrow$ 4.39\% (DC+Thresh) --- improved by 0.98pp but still $<$5\%
\end{itemize}

\paragraph{Analysis: Why Thresholding Fails.} Confidence distribution analysis (from \texttt{refusal\_confidence\_analysis.ipynb}) reveals the fundamental limitation:

\begin{enumerate}
    \item \textbf{Models are more confident when refusing than when correct.} For Qwen3-32B: mean confidence for refusals is 12.17 (lower = more confident) compared to 10.3 for correct answers. This means confidence thresholding that targets refusals will \textit{also} filter out correct answers.
    
    \item \textbf{Refusal confidence is not discriminative.} The confidence distributions for ``correct refusal'' (question genuinely unanswerable) vs ``incorrect refusal'' (question is answerable but model refuses) overlap substantially. Simple thresholds cannot distinguish these cases.
    
    \item \textbf{Model-specific calibration is brittle.} DeepSeek-8B and Qwen3-32B exhibit opposite confidence patterns: DeepSeek shows ``cautious refusal'' (high uncertainty when refusing) while Qwen3 shows ``over-confident refusal'' (high confidence when refusing). Hand-tuned thresholds for one model do not transfer.
\end{enumerate}

\paragraph{Implication for CoRefine.} This analysis motivates the learned controller approach: rather than hand-crafting confidence thresholds, CoRefine trains a neural network to recognize patterns in full confidence traces that distinguish recoverable vs genuine uncertainty. The 76.8\% validation accuracy (vs $<$5\% naive threshold) demonstrates that these patterns exist and are learnable, even if they cannot be captured by simple threshold rules.

\subsection{BixBench Case Studies}
\label{app:bixbench-cases}

We present two additional case studies demonstrating the 4-class controller's behavior on BixBench questions.

\subsubsection{Case Study: Q19 - Successful Refinement with Unsure Exclusion}

\begin{figure}[h]
\centering
\includegraphics[width=\textwidth]{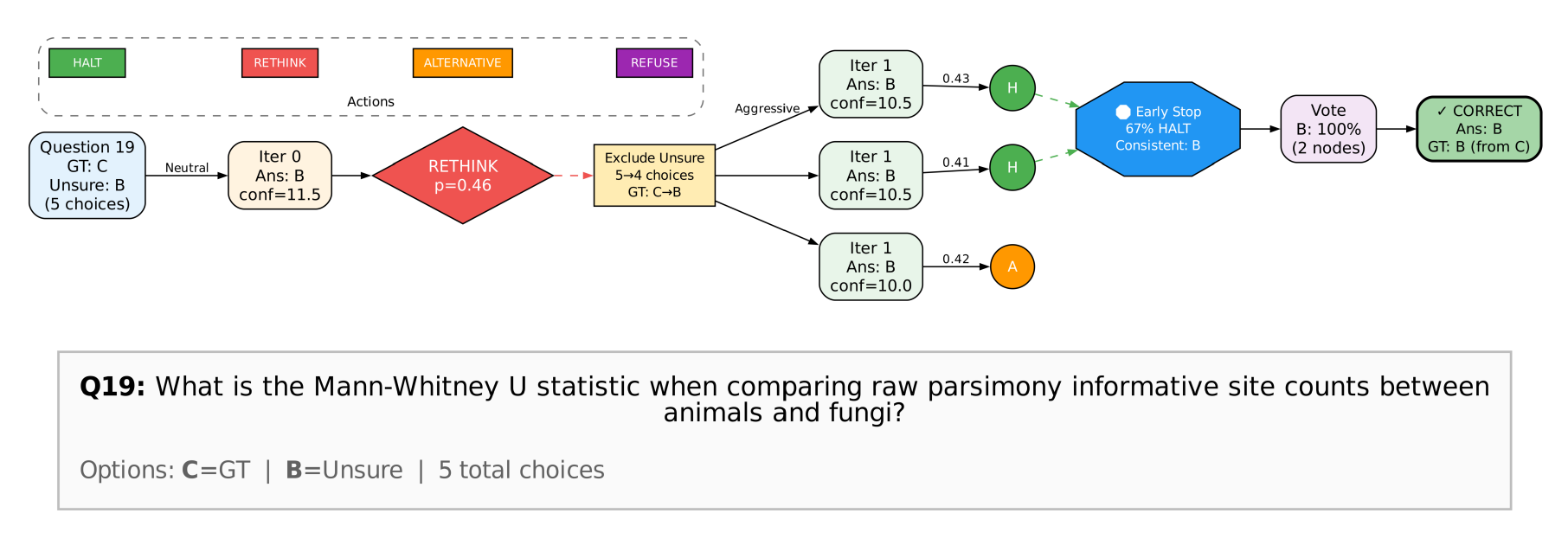}
\caption{\textbf{CoRefine Tree on BixBench Q19} (Mann-Whitney U statistic, ground truth: B after Unsure exclusion). Demonstrates the full refinement pipeline: warmup $\rightarrow$ RETHINK $\rightarrow$ refinement $\rightarrow$ HALT. The controller correctly identifies that the initial moderate-confidence answer warrants verification, then halts after refinement confirms consistent answer B. Node colors: green=HALT, red=RETHINK, orange=ALTERNATIVE.}
\label{fig:bixbench_q19}
\end{figure}

Figure~\ref{fig:bixbench_q19} shows the controller's behavior on a Mann-Whitney U statistic question. The warmup trace selects answer B with confidence 11.49, receiving RETHINK. After refinement, 2 of 3 nodes HALT on B (67\%), with one receiving ALTERNATIVE. The early stopping condition is satisfied with consistent answer B.

\paragraph{Key Observation.} This example demonstrates the ``Unsure Exclusion'' mechanism where ground truth changes from C$\rightarrow$B after removing the ``unsure'' option. The controller successfully navigates this by recognizing that initial uncertainty (RETHINK) can be resolved through additional reasoning.

\subsubsection{Case Study: Q5 - Honest Refusal on Genuine Uncertainty}

\begin{figure}[h]
\centering
\includegraphics[width=\textwidth]{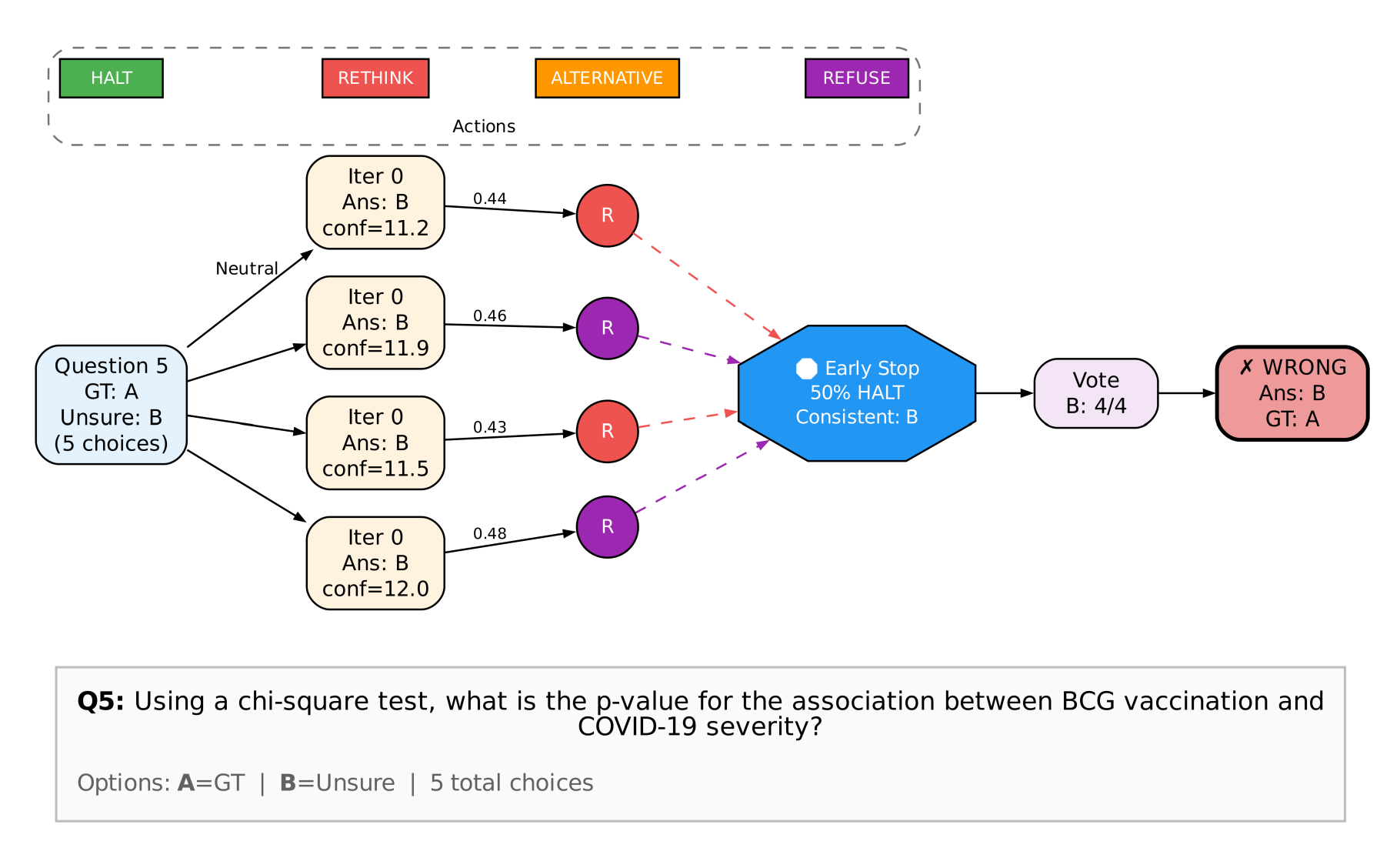}
\caption{\textbf{CoRefine Tree on BixBench Q5} (Chi-square p-value, ground truth: A). The controller recognizes \textit{genuine} uncertainty: all 4 warmup traces select ``Unsure'' with high confidence in that choice. Unlike Q3 where refusal stemmed from post-trained conservatism (3.4\% accuracy recoverable to correct answers), here the 50\% REFUSE rate indicates the controller predicts additional reasoning will not resolve the knowledge gap. Node colors: green=HALT, red=RETHINK, orange=ALTERNATIVE, purple=REFUSE.}
\label{fig:bixbench_q5}
\end{figure}

Figure~\ref{fig:bixbench_q5} shows controller behavior on a chi-square test p-value question where the model exhibits \textit{genuine} uncertainty. All 4 warmup traces select option B (``Insufficient information''), with the controller assigning 2 RETHINK and 2 REFUSE actions (50\% early stop threshold met).

\paragraph{Key Observation.} This case demonstrates \textit{honest refusal}---a critical capability for regulated domains. The controller distinguishes between: (1) over-trained conservative refusal that can be overcome with encouragement (Q3: all RETHINK $\rightarrow$ successful refinement), and (2) genuine knowledge limitations warranting abstention (Q5: 50\% REFUSE $\rightarrow$ early stop). This distinction explains the dramatic accuracy difference: baseline methods drop from 38.5\% to 3.4\% because they cannot differentiate these cases, while CoRefine's 4-class controller learns when encouragement will succeed versus when honest refusal is appropriate.

\subsubsection{Summary: 4-Class Controller Behavior}

The three BixBench case studies (Q3 in Section~\ref{sec:bixbench}, Q19 and Q5 above) demonstrate the 4-class controller's key capability:

\begin{table}[h]
\centering
\caption{BixBench case studies: 4-class controller distinguishes refusal types.}
\small
\begin{tabular}{lcccl}
\toprule
\textbf{Question} & \textbf{Warmup Behavior} & \textbf{Controller Action} & \textbf{Result} & \textbf{Interpretation} \\
\midrule
Q3 (Odds ratio) & 3/4 Unsure & 4/4 RETHINK & Correct (D) & Recoverable conservatism \\
Q19 (Mann-Whitney) & 1/1 B (uncertain) & RETHINK & Correct (B) & Verification succeeds \\
Q5 (Chi-square) & 4/4 Unsure & 2 RETHINK + 2 REFUSE & Abstain & Genuine uncertainty \\
\bottomrule
\end{tabular}
\end{table}

The controller's ability to predict \textit{when refinement will succeed} is the key to recovering accuracy from the 3.4\% baseline. By learning confidence patterns that distinguish post-trained humility from genuine knowledge gaps, CoRefine enables selective encouragement rather than blanket aggressive prompting.

\subsection{Key Takeaways}

\begin{enumerate}
    \item \textbf{Dual-OOD Challenge:} Regulated domains require adaptation to both knowledge distribution (biology vs. mathematics) and behavioral distribution (refusal vs. mandatory answering).
    \item \textbf{Over-Refusal Problem:} Safety-tuned models default to abstention even when additional reasoning could resolve uncertainty (38.5\% → 3.4\% accuracy drop).
    \item \textbf{Lightweight Adaptation:} A 42K-parameter controller learns when to accept refusal vs. push for answers, avoiding expensive full fine-tuning.
    \item \textbf{Prompt Distribution Fidelity:} Two-phase prompting (neutral → aggressive) preserves training distribution while enabling refinement.
    \item \textbf{Model-Specific Patterns:} Refusal confidence patterns vary by model architecture (Qwen3: over-confident, DeepSeek: cautious), requiring calibrated thresholds.
\end{enumerate}

\section{Synthesis Prompt Templates}
\label{app:prompts}

\subsection{RETHINK Prompt}

\begin{tcolorbox}[colback=gray!5, colframe=gray!50, title=RETHINK Synthesis Prompt]
\small
You are solving a mathematical problem. Your previous attempt may have errors.

\textbf{Problem:} \{problem\}

\textbf{Previous Attempts:}
\{compacted\_history\}

\textbf{Previous Answer:} \{previous\_answer\}
\textbf{Confidence:} \{confidence\_stats\}

\textbf{Task:} Re-examine your reasoning step by step. Verify each calculation and logical inference. Consider whether your approach is sound. If you find errors, correct them. If your reasoning is correct, confirm your answer.

Please provide your solution with final answer in $\backslash$boxed\{\}.
\end{tcolorbox}

\subsection{ALTERNATIVE Prompt}

\begin{tcolorbox}[colback=gray!5, colframe=gray!50, title=ALTERNATIVE Synthesis Prompt]
\small
You are solving a mathematical problem. Your previous approaches have not succeeded.

\textbf{Problem:} \{problem\}

\textbf{Previous Attempts:}
\{compacted\_history\}

\textbf{Task:} Your previous approaches may have fundamental issues. Try a COMPLETELY DIFFERENT method or problem formulation. Consider:
- Alternative problem representations
- Different mathematical techniques
- Unconventional solution paths

Please provide your solution with final answer in $\backslash$boxed\{\}.
\end{tcolorbox}

\section{Related Work (Extended)}
\label{app:related-work}

\subsection{Test-Time Scaling}
Current LLMs increasingly succeed by allocating very large amounts of reasoning at inference, a paradigm we call test-time scaling~\citep{snell2024scaling,welleck2024decoding}. Along one axis, Chain-of-Thought~\citep{wei2022chain} depth is scaled by lengthening a single reasoning trajectory through more thinking steps; representative models include o1~\citep{jaech2024openai}, DeepSeek R1~\citep{guo2025deepseek}, Kimi K1.5~\citep{team2025kimi}, Qwen3~\citep{yang2025qwen3}, and Grok-4~\citep{xai2025grok4}. Along a complementary axis, parallel generation is scaled by increasing the number of trajectories and aggregating them: Self-Consistency~\citep{Wang2023SelfConsistency} and Best-of-N~\citep{brown2024large,irvine2023rewarding} sample multiple candidates and select via voting or a score. CoRefine introduces a third axis: sequential refinement with learned halting.

\subsection{Efficient Reasoning}
Test-time scaling for reasoning seeks better accuracy-compute trade-offs through adaptive sampling and richer aggregation. On the parallel axis, Early-Stopping Self-Consistency (ESC), Adaptive-Consistency, Dynamic Voting, and Dynasor achieve more efficient self-consistency by reducing the required sample count while preserving accuracy~\citep{li2024escape,aggarwal2023let,xue2023dynamic,fu2024efficiently}. On the CoT-depth axis, efficient CoT fine-tuning methods elicit shorter, more efficient chains~\citep{chen2024not,luo2025o1,hou2025thinkprune}. CoRefine contributes a sequential refinement approach that complements both axes.

\subsection{Confidence Estimation}
Confidence estimation techniques offer a complementary direction by directly quantifying the reliability of model outputs. DeepConf~\citep{fu2025deepconf} demonstrates that confidence-filtered majority voting substantially outperforms naive self-consistency, establishing that token-level confidence provides discriminative signal for solution quality. Related work proposes metrics such as token-level entropy and uncertainty scores~\citep{Fadeeva2024FactChecking}, self-certainty based on KL divergence from a uniform distribution~\citep{kang_scalable_2025}, and specialized confidence tokens learned during fine-tuning~\citep{chuang2025learning,Zhao2025Learning}. The ART framework~\citep{Shridhar2023ART} introduced trust scoring for iterative refinement, while path-consistency methods~\citep{Zhu2024PathConsistency} leverage high-confidence partial reasoning prefixes to guide subsequent sampling. Auxiliary lightweight predictors such as UHeads~\citep{Ni2025UHead} provide task-agnostic step-level uncertainty estimates. CoRefine uniquely uses full-trace confidence as a control signal for refinement decisions rather than for trace filtering or ranking.

\subsection{Self-Refinement and Error Correction}
Self-refinement methods enable LLMs to iteratively improve their outputs through feedback loops~\citep{madaan2023self}. However, recent work reveals fundamental limitations: \citet{Seo2024Rethinking} demonstrate that refined code is not always superior to original versions, motivating learned stopping criteria rather than fixed iteration counts. The AutoCrit framework~\citep{Sang2025AutoCrit} introduces meta-reasoning with dedicated critique agents and execution monitors, achieving 12--18\% accuracy improvements by catching mistakes ``in the moment.'' This approach requires expensive additional model calls; CoRefine achieves similar benefits through lightweight confidence-based control. 

Recent confidence-guided refinement methods provide complementary perspectives: ConCISE~\citep{Qiao2025ConCISE} monitors step-wise confidence for deficits, triggering early stopping or confidence phrase insertion to achieve $\sim$50\% token-length savings. CISC~\citep{Taubenfeld2025CISC} uses softmax-normalized confidence for weighted voting, reducing sampling costs by 40\% while preserving accuracy. C2R~\citep{Jang2025C2R} curates diverse sub-question chains and selects those with sufficiently high confidence margins for zero-shot QA. Cost-effective refinement remains an active area: CERET~\citep{Cai2024CERET} provides extrinsic refinement using semantic stability and entailment without iterative LLM inference, while \citet{Tang2024REx} frame code repair as an exploration-exploitation tradeoff via Thompson Sampling. CoRefine's RETHINK vs. ALTERNATIVE decision space embodies this tradeoff with learned confidence-based routing.

\subsection{Error Propagation in Reasoning}
A critical challenge in multi-step reasoning is error propagation---the tendency for early mistakes to amplify downstream~\citep{Sang2025AutoCrit}. \citet{Feng2025Misinfo} study misinformation propagation in LLM reasoning chains, finding that models fail to correct errors over half the time and that \emph{early factual corrections} are the most effective mitigation. This motivates CoRefine's design: confidence-guided intervention enables early detection and correction before errors compound. The FAIR-RAG framework~\citep{Aghajani2025FAIRRAG} addresses related issues through structured evidence assessment gating, while AgentErrorBench~\citep{Liang2025COCO} demonstrates that systematic learning from failures can improve agent success rates by 26\%. CoRefine's controller learns similar patterns from historical trajectories, enabling proactive intervention based on confidence signals.

\subsection{Context Engineering for Refinement}
Effective refinement requires managing context across iterations. \citet{Anthropic2025Context} identify context engineering as essential for AI agents, noting that LLMs have finite attention budgets and suffer from ``context rot'' as token counts increase. Key strategies include compaction, summarization, and treating the file system as unlimited context~\citep{Thatipalli2025Context}. CoRefine's synthesis prompts implement these principles: previous reasoning attempts are compacted into high-signal summaries that extract key insights, identified errors, and promising directions, avoiding the ``lost in the middle'' problem while preserving actionable information for subsequent iterations.

\section{Failure Analysis}
\label{app:failures}

\subsection{Common Failure Modes}

\paragraph{False HALT.} The controller occasionally halts prematurely when an incorrect answer has high confidence (overconfident wrong answer). This occurs in approximately 5\% of cases.

\paragraph{Excessive iteration.} On some problems, the controller fails to converge to HALT, reaching the maximum iteration limit. This typically occurs when confidence oscillates between moderate values.

\paragraph{Answer extraction errors.} Complex mathematical expressions with nested braces can cause answer extraction failures, leading to false ``inconsistent answer'' signals.

\subsection{Mitigation Strategies}

\begin{itemize}
    \item \textbf{CoRefine Tree (Hybrid mode):} Use branching refinement with $K=4$ warmup traces and branch factor $B=2$, with early stopping when halt rate $>$50\%. This provides robustness while maintaining token efficiency.
    \item \textbf{Answer consistency override:} If same answer appears 3+ times, force HALT regardless of confidence
    \item \textbf{Maximum iteration cap:} Enforce reasonable upper bound (20 iterations) to prevent runaway computation
\end{itemize}

\section{Reproducibility}
\label{app:reproducibility}



\subsection{Dataset Information}

All evaluation datasets are publicly available:
\begin{itemize}
    \item AIME 2024/2025: \url{https://artofproblemsolving.com/wiki/index.php/AIME}
    \item HMMT 2025: \url{https://hmmt.org}
    \item BRUMO 2025: \url{https://brumo.org}
\end{itemize}

\subsection{Compute Requirements}

\begin{itemize}
    \item Controller training: <1 GPU-hour on A100
    \item Full benchmark evaluation: $\sim$24 GPU-hours
    \item Minimum requirements: Single GPU with 24GB VRAM (with model quantization)
\end{itemize}

\end{document}